\pdfoutput=1

\documentclass{article}

\PassOptionsToPackage{numbers}{natbib}

\usepackage[preprint]{neurips_2025}

\usepackage[utf8]{inputenc} % allow utf-8 input
\usepackage[T1]{fontenc}    % use 8-bit T1 fonts
\usepackage{hyperref}       % hyperlinks
\usepackage{url}            % simple URL typesetting
\usepackage{booktabs}       % professional-quality tables
\usepackage{amsfonts}       % blackboard math symbols
\usepackage{nicefrac}       % compact symbols for 1/2, etc.
\usepackage{microtype}      % microtypography
\usepackage{xcolor}         % colors
\usepackage{graphicx}
\usepackage{amsmath}
\usepackage{multirow}
\usepackage{tabularx}
\usepackage[ruled,vlined,linesnumbered,ruled]{algorithm2e}
\usepackage{listings}
\usepackage{xcolor}
\usepackage{array}
\usepackage{adjustbox}
\usepackage{ragged2e}

\usepackage{tabularray}
\UseTblrLibrary{graphicx}

\newcommand{\imgcenter}[3][2cm]{%
	\vspace{0.5em}%
	\adjustbox{valign=m}{\includegraphics[width=#1, bb=#3]{#2}}%
	\vspace{0.5em}%
}

\definecolor{codeblue}{rgb}{0.25,0.5,0.75}
\definecolor{codegray}{rgb}{0.4,0.4,0.4}
\definecolor{codegreen}{rgb}{0,0.6,0}
\definecolor{backcolor}{rgb}{0.95,0.95,0.92}

\lstdefinestyle{mystyle}{
    language=Python,
    backgroundcolor=\color{backcolor},   
    commentstyle=\color{codegreen},
    keywordstyle=\color{codeblue}\bfseries,
    numberstyle=\tiny\color{codegray},
    stringstyle=\color{orange},
    basicstyle=\ttfamily\footnotesize,
    breakatwhitespace=false,         
    breaklines=true,                 
    captionpos=b,                    
    keepspaces=true,                 
    numbers=none,                    
    numbersep=5pt,                  
    showspaces=false,                
    showstringspaces=false,
    showtabs=false,                  
    tabsize=4
}

\title{Agent-Centric Personalized Multiple Clustering with Multi-Modal LLMs}

\author{
	Ziye Chen\textsuperscript{1,2}  \ \ \ \
    Yiqun Duan\textsuperscript{1}\thanks{Co-corresponding author. This work was completed by Ziye Chen during his internship at TikTok, Australia.}  \ \ \ \
    Riheng Zhu\textsuperscript{1}  \ \ \ \
    Zhenbang Sun\textsuperscript{1}  \ \ \ \
	Mingming Gong\textsuperscript{2}\footnotemark[1] \\
	\textsuperscript{1} TikTok, Australia \\
    \textsuperscript{2} School of Mathematics and Statistics, University of Melbourne, Australia \\
	\tt\small ziyec1@student.unimelb.edu.au, \tt\small yiqunduan@bytedance.com, \tt\small zhuriheng@bytedance.com, \\ \tt\small sunzhenbang@bytedance.com, \tt\small mingming.gong@unimelb.edu.au
}

\begin{document}

\maketitle

\begin{abstract}
Personalized multiple clustering aims to generate diverse partitions of a dataset based on different user-specified aspects, rather than a single clustering. It has recently drawn research interest for accommodating varying user clustering preferences.
Recent approaches primarily use CLIP embeddings with proxy learning to extract representations biased toward user interests. However, CLIP primarily focuses on coarse image-text alignment, lacking a deep contextual understanding of user interests.
To overcome these limitations, we propose an agent-centric personalized clustering framework that leverages multi-modal large language models (MLLMs) as agents to comprehensively traverse a relational graph to search for clusters based on user interests. Due to the advanced reasoning mechanism of MLLMs, the obtained clusters align more closely with user-defined criteria than those obtained from CLIP-based representations. To reduce computational overhead, we shorten the agents' traversal path by constructing a relational graph using user-interest-biased embeddings extracted by MLLMs. 
A large number of weak edges can be filtered out based on embedding similarity, facilitating an efficient traversal search for agents. Experimental results show that the proposed method achieves NMI scores of $0.9667$ and $0.9481$ on the Card Order and Card Suits benchmarks, respectively, largely improving the SOTA model by over $140\%$.
\end{abstract}

\section{Introduction}
\label{sec:intro}
Clustering is a fundamental technique that partitions data into meaningful groups, uncovering underlying structures within a dataset. Traditional clustering methods ~\cite{ng2001spectral,bishop2006pattern,caron2018deep,caron2020unsupervised} rely on handcrafted features or fixed representations, which may fail to capture the inherent complex data relationships. Recent advances \cite{lochman2024learned,chu2024image,qian2023stable,ouldnoughi2023clip,qian2022unsupervised,li2021contrastive,duan2025scaling} leverage learning-based techniques to obtain more expressive representations, leading to significant performance improvements. Most of these approaches produce only one data partition from a single perspective.

\begin{figure}[t]
\centering
\includegraphics[bb=6 8 1272 588, width=1.0\linewidth]{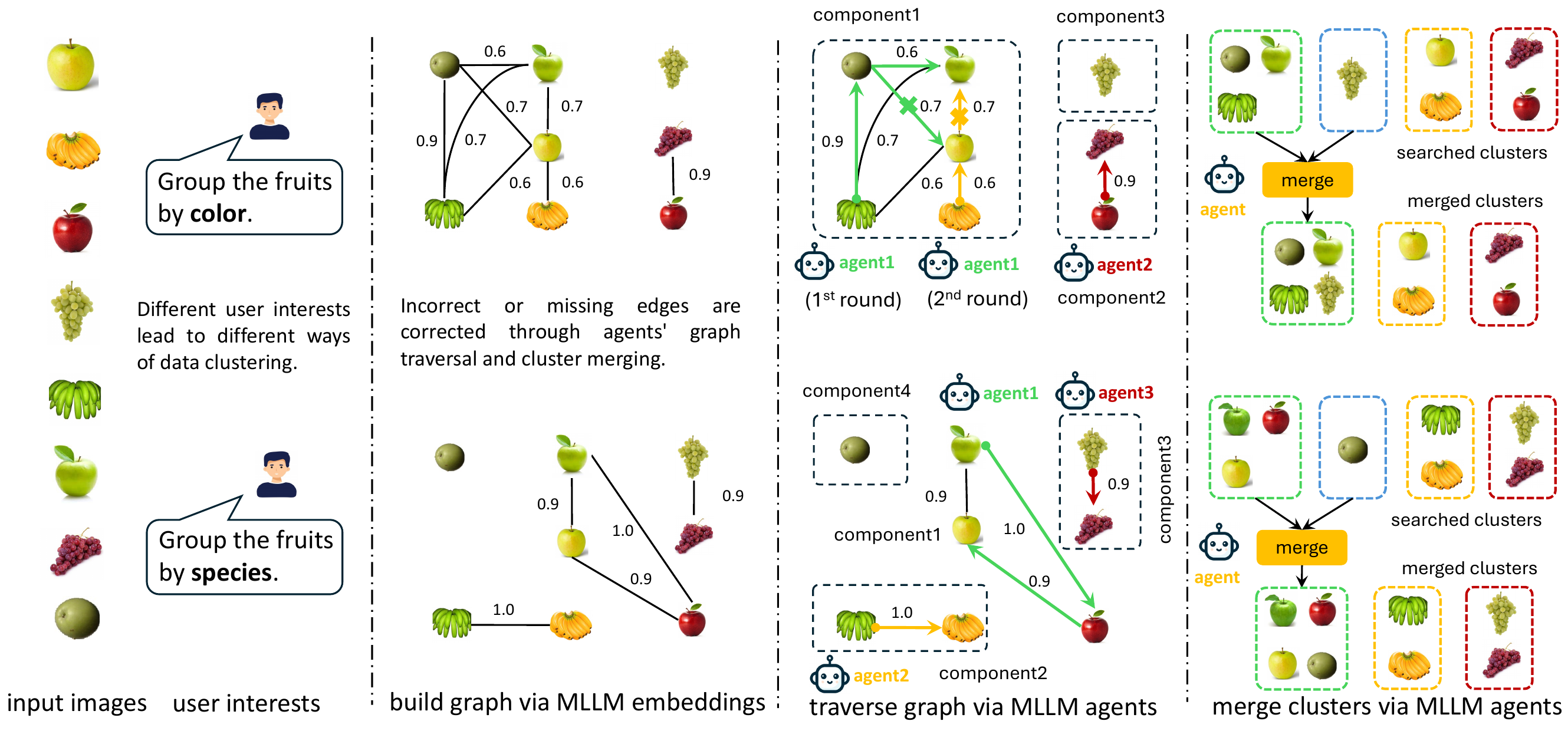}
\caption{The workflow of the proposed agent-centric multiple clustering framework, which obtains a personalized clustering by using MLLMs as agents to traverse a relational graph based on user preferences. The relational graph is constructed from MLLM embeddings biased toward user interests.}
\label{Fig:illustration}
% \vspace{-3mm}
\end{figure}

However, real-world data exhibit inherent complexity, making it impossible for any single clustering method to capture all relevant structures. Consequently, identifying multiple valid clusterings within a dataset is essential to address various analytical needs. For example, the fruits in Fig. \ref{Fig:illustration} can be grouped by color or species, depending on the perspective. This need has driven the development of personalized multiple clustering, which aims to generate diverse partitions of a dataset based on different user clustering preferences, rather than a single fixed solution.

Typical multiple clustering methods \cite{miklautz2020deep,ren2022diversified,yao2023augdmc} primarily leverage autoencoders and data augmentation to extract diverse data representations, enabling clustering from different perspectives. However, identifying which clustering outcome aligns with user interests remains challenging, as these interests are often expressed through abstract keywords—such as color or species in the case of fruits, as illustrated in Fig. \ref{Fig:illustration}. Recent approaches \cite{yao2024multi,yao2024customized} integrate CLIP \cite{radford2021learning,li2022exploring,li2023clip,li2025closer} embeddings with proxy learning to obtain data representations tailored to user clustering preferences. While effective, CLIP is trained on coarse-grained image-text pairs and is designed for overall cross-modal alignment, limiting its ability to focus solely on fine-grained, user-specified aspects.

To address these limitations, we propose an \textbf{agent-centric personalized multiple clustering framework} that employs shared multi-modal large language models (MLLMs) as agents to assess data relationships based on user interests and search for clusters accordingly. As shown in Fig.~\ref{Fig:illustration}, we first construct a relational graph where nodes represent input images and edges encode pairwise similarities. Each agent is assigned to a connected component of the graph and initializes a cluster from the highest-degree node within that component. It then expands the cluster by traversing neighboring nodes, evaluating their membership based on user interests, and updating the cluster boundaries accordingly. Once a cluster is completed, the agent selects the next highest-degree unassigned node in the component to start a new cluster, repeating the traversal process until all nodes are assigned. After all agents complete, another agent performs a global review to merge semantically redundant clusters. Errors in graph connectivity are corrected through agent-based graph traversal and cluster merging. Leveraging the reasoning capabilities of MLLMs, our method yields clusters more faithfully aligned with user-defined criteria than those from CLIP-based approaches.

A naive approach to constructing the relational graph is to fully connect all nodes, resulting in an overly dense graph with many spurious edges. This significantly increases the traversal steps needed for agents to discard falsely connected neighbors, thereby compromising clustering efficiency. To address this, we build a weighted graph where edge weights reflect similarities between data embeddings extracted by MLLMs conditioned on user interests. By pruning weak edges based on similarity scores, we retain a compact set of high-quality neighbors, substantially reducing the traversal steps required for agents to evaluate them and improving clustering efficiency. Specifically, for each input image, the MLLM generates a textual description containing an embedding token based on user interests, where the hidden state of this token is projected as the data embedding. Unlike CLIP-based embeddings, which are limited by coarse-grained modality alignment, MLLM-based embeddings adaptively capture fine-grained, user-specified semantics through contextual reasoning, resulting in a sparser relational graph with primarily meaningful edges.

We evaluate our method on five publicly available visual datasets commonly used for multiple clustering tasks, and the results demonstrate its state-of-the-art performance. For instance, our method achieves NMI scores of $0.9667$ and $0.9481$ on the Card Order and Card Suits benchmarks, respectively, \textbf{surpassing the current state-of-the-art model by over $140\%$}. To the best of our knowledge, this is the first multiple clustering method built on MLLMs, providing a promising approach to unify personalized clustering and agent-based searching.

\section{Related Work}
\label{sec:related}

\paragraph{Multiple Clustering} Multiple clustering explores diverse data partitions from different perspectives, gaining increasing attention. Early methods rely on hand-crafted rules and representations. For example, COALA \cite{bae2006coala} generates new clusters using existing ones as a constraint, Hu et al. \cite{hu2017finding} maximized eigengap across subspaces, and Dang et al. \cite{dang2010generation} utilizes an expectation-maximization framework to optimize mutual information. Recent approaches leverage learning-based techniques for better representations. For instance, ENRC \cite{miklautz2020deep} optimizes clustering objectives within a latent space learned by an auto-encoder, iMClusts \cite{ren2022diversified} leverages auto-encoders and multi-head attention to learn diverse feature representations, and AugDMC \cite{yao2023augdmc} applies data augmentation to generate diverse image perspectives. However, it remains challenging to identify the clustering most relevant to user interests. Recently, Multi-MaP \cite{yao2024multi} and Multi-Sub \cite{yao2024customized} integrate CLIP embeddings with proxy learning to generate data representations aligned with user interests. While effective, CLIP lacks the deep contextual understanding necessary to capture abstract and nuanced user interests.

\paragraph{LLM-Driven Agent Search} LLM-driven agent search integrates Large Language Models (LLMs) into search processes, enabling agents to reason, plan, and interact with tools for efficient information retrieval. For example, PaSa \cite{he2025pasa} employs LLM-driven agents to search, read papers, and select references, ensuring comprehensive and accurate results for scholarly queries. Toolformer \cite{schick2024toolformer} demonstrates how LLMs can interact with external APIs (e.g., search engines, calculators) to enhance factual accuracy. ReAct \cite{yao2022react} combines reasoning and acting, allowing agents to iteratively retrieve and process information for more effective decision-making. Recently, this field has shifted toward multi-agent search, emphasizing cooperative problem-solving, knowledge sharing, and decentralized decision-making. For example, CAMEL \cite{li2023camel} introduces LLM agents with specialized roles (e.g., teacher and student) to refine search strategies collaboratively. Voyager \cite{wang2023voyager} applies LLMs in an open-world exploration setting (Minecraft), where agents autonomously collect, analyze, and apply new knowledge. In this paper, we apply MLLM-driven agents to search clusters based on user clustering preferences.

\section{Method}
\label{sec:method}
In this section, we introduce our agent-centric personalized multiple clustering framework. We begin with an overview of the framework, followed by a detailed description of each component: agent-centric graph traversal, MLLM-based graph construction, and agent-based membership assessment.

\begin{figure*}[t]
\centering
\includegraphics[bb=0 4 1096 596, width=1.0\linewidth]{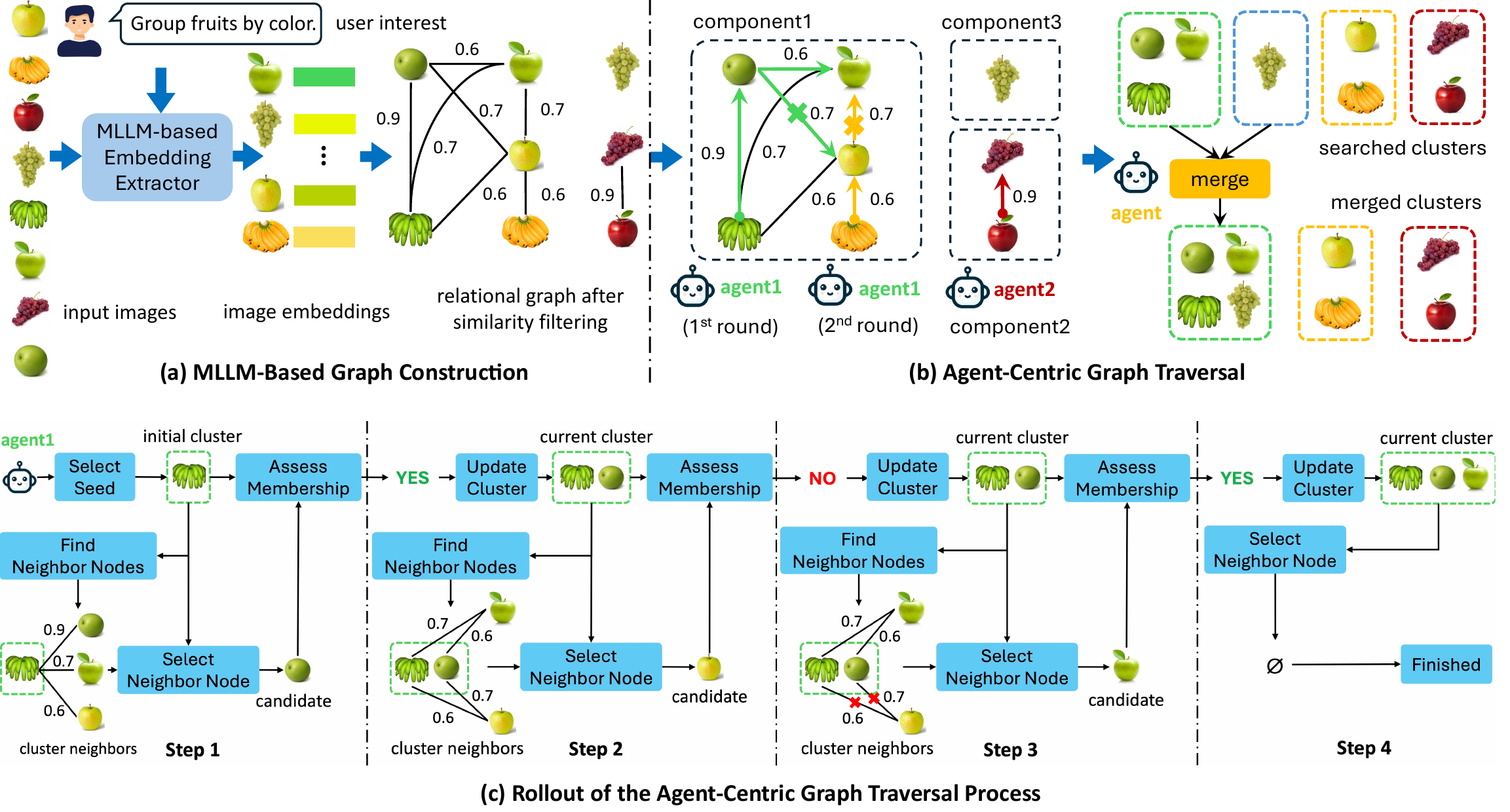}
\caption{Overview of the Agent-Centric Personalized Multiple Clustering Framework. (a) MLLM-based graph construction, where image embeddings are extracted using MLLM based on user interests, from which a relational graph is constructed. (b) Agent-centric graph traversal, where agents search for clusters by traversing the graph. (c) Rollout of the graph traversal process, where agents expand cluster iteratively by assessing neighboring nodes based on user-defined criteria.}
\label{Fig:framework}
% \vspace{-1mm}
\end{figure*}

\subsection{Overall Framework}
The framework, illustrated in Fig. \ref{Fig:framework}, begins by extracting image embeddings using an MLLM-based embedding extractor, tailoring representations to user interests. A relational graph is then constructed from these embeddings, where a large number of weak edges are filtered out based on embedding similarities. Next, multiple MLLM-based agents traverse this graph to search for clusters aligned with user preferences, with each agent specializing in a different connected component of the graph. After all agents complete, another agent performs a global review to merge semantically redundant clusters. This agent-centric approach produces clusters that better adhere to user-defined criteria than those obtained from CLIP-based representations.

\subsection{Agent-Centric Graph Traversal}
\label{sec:agent_traversal}
Here, we describe our agent-centric graph traversal approach for cluster discovery. We begin with a relational graph $\mathcal{G}=\{\mathcal{V}, \mathcal{E}, \mathcal{W}\}$ constructed as detailed in Sec~\ref{sec:graph_construction}, where each node $u \in \mathcal{V}$ represents an input image, and each edge $\{u, v\} \in \mathcal{E}$ carries a weight $w(u, v) \in \mathcal{W}$ indicating a precomputed, embedding-based similarity score between $u$ and $v$. Leveraging shared multi-modal large language models (MLLMs) as agents, we traverse $\mathcal{G}$ to search for clusters $\mathcal{S}$ aligned with user interests $T$.

As outlined in Algorithm \ref{alg:agent_traversal}, we first identify the connected components of the relational graph $\mathcal{G}$, resulting in a total of $M$ components:
\begin{equation}
    \{\mathcal{C}_i\}_{i=1}^M = \text{ConnectedComponents}(\mathcal{G}).
    \label{Eq:1}
\end{equation}
For each connected component $\mathcal{C}_i$, we select the highest-degree node $v_i^*$ and designate it as the seed for an initial cluster $\mathcal{S}_i$ as follows:
\begin{equation}
    v_i^* = \arg\max_{v \in \mathcal{C}_i} \sum_{u \in \mathcal{N}(v)} w(v, u), \quad \mathcal{S}_i = \big\{ v_i^* \big\},
    \label{Eq:2}
\end{equation}
where $\mathcal{N}(v)$ is the neighborhood of node $v$. We then assign an MLLM-based agent $\mathcal{A}_i$ to each initial cluster $\mathcal{S}_i$, resulting in a total of $M$ agents, denoted as $\{ \mathcal{A}_i \}_{i=1}^M$. Each agent $\mathcal{A}_i$ iteratively expands its corresponding cluster $\mathcal{S}_i$ by evaluating the membership of its neighboring nodes in $\mathcal{S}_i$ based on user interests $T$. As shown in Fig. \ref{Fig:framework} (c) and Algorithm 1, at each step, the agent $\mathcal{A}_i$ first determines the neighborhood of the current cluster $\mathcal{S}_i$, defined as:
\begin{equation}
    % \mathcal{N}(\mathcal{S}_i) = \big\{ u \in \mathcal{V} \setminus \mathcal{S}_i \mid \exists v \in \mathcal{S}_i, (u, v) \in \mathcal{E} \big\}.
    \mathcal{N}(\mathcal{S}_i) = \bigcup_{v \in \mathcal{S}_i} \mathcal{N}(v) \setminus \mathcal{S}_i.
    \label{Eq:3}
\end{equation}
Next, the agent selects a candidate neighboring node $v_j^*$	from $\mathcal{N}(\mathcal{S}_i)$ that exhibits the highest weighted connectivity to $\mathcal{S}_i$, given by:
\begin{equation}
    v_j^* = \arg\max_{v \in \mathcal{N}(\mathcal{S}_i)} \sum_{u \in \mathcal{S}_i} w(v, u).
    \label{Eq:4}
\end{equation}
The agent then assesses whether $v_j^*$ should be included in $\mathcal{S}_i$ based on user-defined criteria $T$, as detailed in Sec. \ref{sec:membership_assessment}. If the candidate node $v_j^*$ is deemed a valid member, it is merged into the current cluster $\mathcal{S}_i$, updating $\mathcal{S}_i$ as follows:
\begin{equation}
    \mathcal{S}_i \gets \mathcal{S}_i \cup \big\{ v_j^* \big\}.
    \label{Eq:5}
\end{equation}
Otherwise, the edges between the candidate node $v_j^*$ and the current cluster $\mathcal{S}_i$ are removed from the edge set $\mathcal{E}$, yielding an updated edge set:
\begin{equation}
    \mathcal{E} \gets \mathcal{E} \setminus \{ (u, v_j^*) \in \mathcal{E} \mid u \in \mathcal{S}_i \}.
    \label{Eq:6}
\end{equation}
The process from Eq. (\ref{Eq:3}) to Eq. (\ref{Eq:6}) is repeated until the neighborhood $\mathcal{N}(\mathcal{S}_i)$ is empty, determining the cluster $\mathcal{S}_i$. The agent then removes the nodes in $\mathcal{S}_i$ from the component $\mathcal{C}_i$ as follows:
\begin{equation}
\mathcal{C}_i \gets \mathcal{C}_i \setminus \mathcal{S}_i.
\label{Eq:7}
\end{equation}
Next, the agent selects the highest-degree unassigned node in $\mathcal{C}_i$ to initiate a new cluster, repeating the traversal process from Eq. (\ref{Eq:2}) to Eq. (\ref{Eq:7}) until all nodes in $\mathcal{C}_i$ are assigned. The $M$ agents, $\{ \mathcal{A}_i \}_{i=1}^M$, traverse the graph $\mathcal{G}$ in parallel, with the searched clusters collected into the set $\mathcal{S}$. Finally, another agent performs a global review to merge semantically redundant clusters in $\mathcal{S}$ based on user interests $T$. Incorrect or missing edges in the graph $\mathcal{G}$ are corrected through agent-based graph traversal and cluster merging. Leveraging the contextual reasoning capabilities of MLLMs, our method yields clusters more faithfully aligned with user interests compared to CLIP-based approaches.

\begin{algorithm}[t]
\footnotesize
\caption{Agent-Centric Graph Traversal for Personalized Multiple Clustering}
\label{alg:agent_traversal}
\DontPrintSemicolon
\KwIn{Relational graph $\mathcal{G} = (\mathcal{V}, \mathcal{E}, \mathcal{W})$, user interests $T$}
\KwOut{Final set of clusters $\mathcal{S}$}

\SetKwFunction{ConnectedComponents}{ConnectedComponents}
\SetKwFunction{Argmax}{argmax}

Initialize $\mathcal{S} \gets \emptyset$\;
$\{\mathcal{C}_1, \mathcal{C}_2, \dots, \mathcal{C}_M\} \gets \ConnectedComponents(\mathcal{G})$\;
\For{$i \gets 1$ \KwTo $M$}{
    \While{$\mathcal{C}_i \neq \emptyset$}{
        $v_i^* \gets \Argmax_{v \in \mathcal{C}_i} \sum_{u \in \mathcal{N}(v)} w(v, u); \quad \mathcal{S}_i \gets \{ v_i^* \}$ \tcp*[r]{Initialize cluster}
        % Assign agent $\mathcal{A}_i$ to cluster $\mathcal{S}_i$\;
        $\mathcal{N}(\mathcal{S}_i) \gets \bigcup_{v \in \mathcal{S}_i} \mathcal{N}(v) \setminus \mathcal{S}_i$\;
        \While{$\mathcal{N}(\mathcal{S}_i) \neq \emptyset$}{
            $v_j^* \gets \Argmax_{v \in \mathcal{N}(\mathcal{S}_i)} \sum_{u \in \mathcal{S}_i} w(v,u)$\;
            \If{agent \(\mathcal{A}_i\) determines \(v_j^*\) belongs to \(\mathcal{S}_i\) based on \(T\)}{
                $\mathcal{S}_i \gets \mathcal{S}_i \cup \{ v_j^* \}$\;
            }
            \Else{
                $\mathcal{E} \gets \mathcal{E} \setminus \{(u, v_j^*) \in \mathcal{E} \mid u \in \mathcal{S}_i\}$\;
            }
            $\mathcal{N}(\mathcal{S}_i) \gets \bigcup_{v \in \mathcal{S}_i} \mathcal{N}(v) \setminus \mathcal{S}_i$\;
        }
        $\mathcal{C}_i \gets \mathcal{C}_i \setminus \mathcal{S}_i; \quad \mathcal{S} \gets \mathcal{S} \cup \{ \mathcal{S}_i \}$ \;
    }
}
\While{$\mathcal{S}$ changed}{
    \ForEach{$(\mathcal{S}_p,\mathcal{S}_q) \in \binom{\mathcal{S}}{2}$}{
        \If{agent $\mathcal{A}_{\text{merge}}$ decides to merge $\mathcal{S}_p$ and $\mathcal{S}_q$ based on $T$}{
            $\mathcal{S}_{pq} \gets \mathcal{S}_p \cup \mathcal{S}_q; \quad \mathcal{S} \gets \bigl(\mathcal{S} \setminus \{\mathcal{S}_p,\mathcal{S}_q\}\bigr) \cup \{\mathcal{S}_{pq}\}; \quad \textbf{break}$\;
        }
    }
}
\Return $\mathcal{S}$

\end{algorithm}

\subsection{MLLM-Based Graph Construction}
\label{sec:graph_construction}
Here, we detail the construction of the relational graph $\mathcal{G} = \{ \mathcal{V}, \mathcal{E}, \mathcal{W} \}$ used for agent traversal. Each node $u \in \mathcal{V}$ corresponds to an input image. Each edge $(u, v) \in \mathcal{E}$ is assigned a weight $w(u,v) \in W$, determined by the similarity between the embeddings $\mathcal{H}(u)$ and $\mathcal{H}(v)$. These embeddings are extracted by the MLLM for nodes $u$ and $v$ based on user interests $T$.

\begin{figure}[t]
\centering
\includegraphics[bb=3 0 1163 398, width=1.0\linewidth]{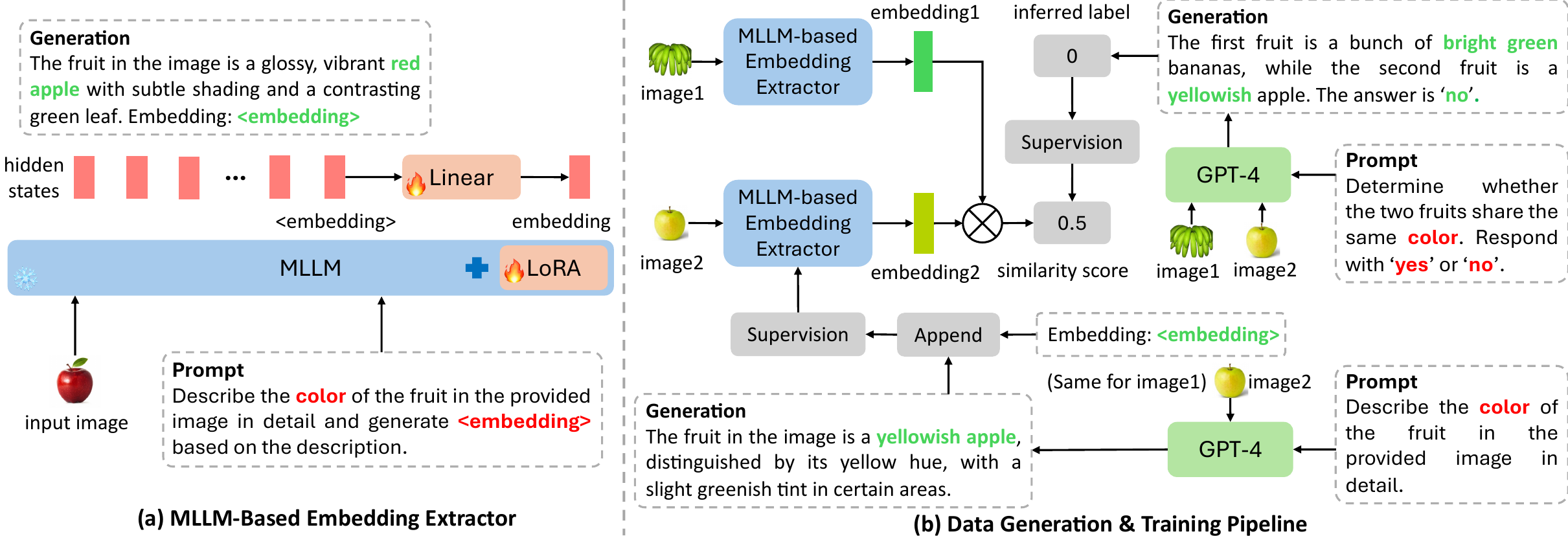}
\caption{Illustration of image embedding extraction using MLLM based on user interests.}
\label{Fig:embedding}
% \vspace{-3mm}
\end{figure}

As shown in Fig. \ref{Fig:embedding} (a), we first design an embedding generation instruction based on user interests $T$. For example, if the user intends to cluster fruit images by ``color'', the instruction is formulated as: ``Describe the color of the fruit in the provided image in detail and generate \texttt{<embedding>} based on the description.'' We then provide the image $u$ and the instruction as inputs to the MLLM, which responds with a reasoning statement followed by an \texttt{<embedding>} token. The hidden state of the \texttt{<embedding>} token is then projected to obtain the user-interest-biased embedding $\mathcal{H}(u)$. After obtaining the embeddings $\mathcal{H}(u)$ and $\mathcal{H}(v)$, the edge weight $w(u, v)$ is computed as:
\begin{equation}
    w(u, v) = \sigma(\beta \cdot \mathcal{H}(u)^{\top} \mathcal{H}(v)),
    \label{Eq:8}
\end{equation}
where $\beta$ is a learnable logit scale, and $\sigma$ denotes the sigmoid function. An edge $(u, v)$ is retained in the edge set $\mathcal{E}$ only if its weight $w(u, v)$ exceeds a predefined threshold $\tau$. 

As shown in Fig. \ref{Fig:embedding} (b), for training data generation, we perform hard negative mining to select the top $1,024$ image pairs with the highest similarity entropy in each epoch. GPT-4 \cite{achiam2023gpt} evaluates the similarity of these pairs based on user interests $T$, from which binary pseudo labels are inferred. It also generates detailed descriptions for each image, with an appended \texttt{<embedding>} token. During training, the MLLM produces descriptions and embeddings for each image pair conditioned on $T$. We compute embedding similarities following Eq. (\ref{Eq:8}) and supervise them using GPT-4’s pseudo labels via binary cross-entropy loss. The generated descriptions are aligned with GPT-4’s outputs using cross-entropy loss. The MLLM is finetuned with LoRA \cite{hu2022lora}.

\begin{figure}[t]
\centering
\includegraphics[bb=0 0 1105 340, width=1.0\linewidth]{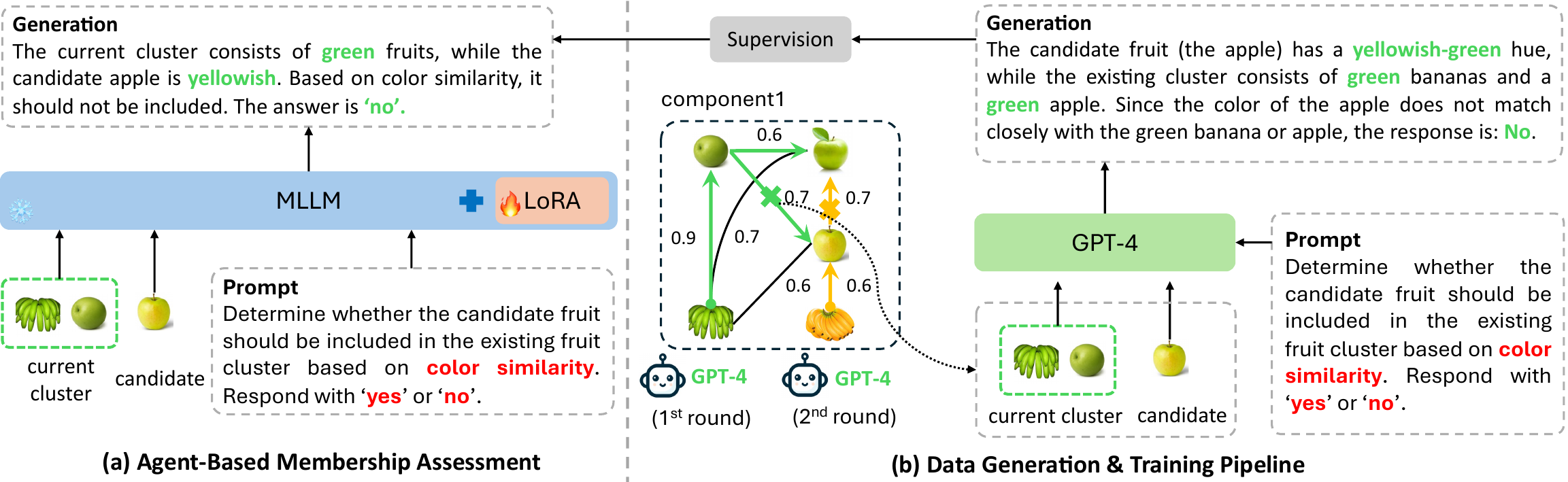}
\caption{Illustration of Agent-based assessment of candidate membership according to user interests.}
\label{Fig:membership}
% \vspace{-3mm}
\end{figure}

\subsection{Agent-Based Membership Assessment}
\label{sec:membership_assessment}
Here, we describe how the agent $\mathcal{A}_i$ assesses whether a candidate node $v_j^*$ should be included in the current cluster $\mathcal{S}_i$ based on user interests $T$. As illustrated in Fig. \ref{Fig:membership} (a), we first select the top-$K$ nodes with the highest degrees within $\mathcal{S}_i$ to form a set of representative nodes:
\begin{equation}
\mathcal{R}_i 
= \arg\max_{\substack{R \subseteq \mathcal{S}_i \\ |\mathcal{R}|=min(K, |\mathcal{S}|)}}
\sum_{v \in R} \sum_{u \in \mathcal{N}(v) \cap \mathcal{S}_i} w(v, u),
\label{Eq:9}
\end{equation}
where $\mathcal{R}_i$ serves as a compact representation of the current cluster $\mathcal{S}_i$. We then design a membership assessment instruction tailored to user interests $T$. For instance, if the user aims to cluster fruit images by ``color'', the instruction is: ``Determine whether the candidate fruit should be included in the existing fruit cluster based on color similarity. Respond with `yes' or `no'.'' The candidate node $v_j^*$ and the representative node set $\mathcal{R}_i$ are provided to the agent $\mathcal{A}_i$ along with this instruction. The agent responds with a reasoning statement followed by a binary decision. Based on the agent's assessment, we update the current cluster $\mathcal{S}_i$ as described in Eqs. (\ref{Eq:5}) and (\ref{Eq:6}).

As shown in Fig. \ref{Fig:membership} (b), for training data generation, we begin by constructing a relational graph using embeddings extracted by the MLLM. We then uniformly sample a subgraph containing $1,024$ nodes from the relational graph. GPT-4 traverses this subgraph according to Eqs. (\ref{Eq:1})–(\ref{Eq:7}), assessing the membership of each neighboring node with respect to the current cluster based on user interests $T$, as defined in Eq. \ref{Eq:9}. We collect these assessments at each traversal step as training samples. During training, the MLLM predicts membership for each sampled neighboring node and its corresponding cluster based on $T$. These predictions are supervised using GPT-4’s assessments via cross-entropy loss. The MLLM is finetuned using LoRA \cite{hu2022lora}.

\section{Experiment}

\begin{table*}[t]
\caption{Comparison with state-of-the-art methods across multiple clustering benchmarks.}
\centering
\footnotesize % 根据需要可改成 \footnotesize 或更小
\resizebox{1.0\textwidth}{!}{
\begin{tabularx}{1.45\textwidth}{l c|c c|c c|c c|c c c c|c c}
\toprule
 & & \multicolumn{2}{c|}{Fruit} & \multicolumn{2}{c|}{Fruit360} & \multicolumn{2}{c|}{Card} & \multicolumn{4}{c|}{CMUface} & \multicolumn{2}{c}{CIFAR10-MC}\\
 & & Color & Species & Color & Species & Order & Suits & Emotion & Sunglass & Identity & Pose & Type & Environment \\
\toprule

\multirow{2}{*}{MSC\cite{hu2017finding}}
 & NMI & 0.6886 & 0.1627 & 0.2544 & 0.2184 & 0.0807 & 0.0497 & 0.1284 & 0.1420 & 0.3892 & 0.3687 & 0.1547 & 0.1136 \\
 & RI  & 0.8051 & 0.6045 & 0.6054 & 0.5805 & 0.7805 & 0.3587 & 0.6736 & 0.5745 & 0.7326 & 0.6322 & 0.3296 & 0.3082 \\
\midrule

\multirow{2}{*}{MCV\cite{guerin2018improving}}
 & NMI & 0.6266 & 0.2733 & 0.3776 & 0.2985 & 0.0792 & 0.0430 & 0.1433 & 0.1201 & 0.4637 & 0.3254 & 0.1618 & 0.1379 \\
 & RI  & 0.7685 & 0.6597 & 0.6791 & 0.6176 & 0.7128 & 0.3638 & 0.5268 & 0.4905 & 0.6247 & 0.6028 & 0.3305 & 0.3344 \\
\midrule

\multirow{2}{*}{ENRC\cite{miklautz2020deep}}
 & NMI & 0.7103 & 0.3187 & 0.4264 & 0.4142 & 0.1225 & 0.0676 & 0.1592 & 0.1493 & 0.5607 & 0.2290 & 0.1826 & 0.1892 \\
 & RI  & 0.8511 & 0.6536 & 0.6868 & 0.6984 & 0.7313 & 0.3801 & 0.6630 & 0.6209 & 0.7635 & 0.5029 & 0.3469 & 0.3599 \\
\midrule

\multirow{2}{*}{iMClusts\cite{ren2022diversified}}
 & NMI & 0.7351 & 0.3029 & 0.4097 & 0.3861 & 0.1144 & 0.0716 & 0.0422 & 0.1929 & 0.5109 & 0.4437 & 0.2040 & 0.1920 \\
 & RI  & 0.8632 & 0.6743 & 0.6841 & 0.6732 & 0.7658 & 0.3715 & 0.5932 & 0.5627 & 0.8260 & 0.6114 & 0.3695 & 0.3664 \\
\midrule

\multirow{2}{*}{AugDMC\cite{yao2023augdmc}}
 & NMI & 0.8517 & 0.3546 & 0.4594 & 0.5139 & 0.1440 & 0.0873 & 0.0161 & 0.1039 & 0.5875 & 0.1320 & 0.2855 & 0.2927 \\
 & RI  & 0.9108 & 0.7399 & 0.7392 & 0.7430 & 0.8267 & 0.4228 & 0.5367 & 0.5361 & 0.8334 & 0.5517 & 0.4516 & 0.4689 \\
\midrule

\multirow{2}{*}{DDMC\cite{yao2024dual}}
 & NMI & 0.8973 & 0.3764 & 0.4981 & 0.5292 & 0.1563 & 0.0933 & 0.1726 & 0.2261 & 0.6360 & 0.4526 & 0.3991 & 0.3782 \\
 & RI  & 0.9383 & 0.7621 & 0.7472 & 0.7703 & 0.8326 & 0.6469 & 0.7593 & 0.7663 & 0.8907 & 0.7904 & 0.5827 & 0.5547 \\
\midrule

\multirow{2}{*}{Multi-Map\cite{yao2024multi}}
 & NMI & 0.8619 & 1.0000 & 0.6239 & 0.5284 & 0.3653 & 0.2734 & 0.1786 & 0.3402 & 0.6625 & 0.4693 & 0.4969 & 0.4598 \\
 & RI  & 0.9526 & 1.0000 & 0.8243 & 0.7582 & 0.8587 & 0.7039 & 0.7105 & 0.7068 & 0.9496 & 0.6624 & 0.7104 & 0.6737 \\
\midrule

\multirow{2}{*}{Multi-Sub\cite{yao2024customized}}
 & NMI & 0.9693 & 1.0000 & 0.6654 & 0.6123 & 0.3921 & 0.3104 & 0.2053 & 0.4870 & 0.7441 & 0.5923 & 0.5271 & 0.4828 \\
 & RI  & 0.9964 & 1.0000 & \textbf{0.8821} & \textbf{0.8504} & 0.8842 & 0.7941 & \textbf{0.8527} & 0.8324 & \textbf{0.9834} & 0.8736 & 0.7394 & 0.7096 \\
\midrule

\multirow{2}{*}{\textbf{Ours}}
 & NMI & \textbf{1.0000} & \textbf{1.0000} & \textbf{0.7214} & \textbf{0.6532} & \textbf{0.9667} & \textbf{0.9481} & \textbf{0.2196} & \textbf{0.9720} & \textbf{0.7631} & \textbf{0.7789} & \textbf{0.6385} & \textbf{0.5931} \\
 & RI  & \textbf{1.0000} & \textbf{1.0000} & 0.8715 & 0.8436 & \textbf{0.9952} & \textbf{0.9882} & 0.8415 & \textbf{0.9936} & 0.9763 & \textbf{0.9002} & \textbf{0.7812} & \textbf{0.7439} \\
\bottomrule
\end{tabularx}
}
% For methods that use k-means for clustering, the algorithm is run 10 times to account for its inherent randomness, with the average clustering performance reported. All methods are evaluated using two metrics: Normalized Mutual Information (NMI) and Rand Index (RI).}
\label{tab:1}
\vspace{-3mm}
\end{table*}

\subsection{Datasets}
\label{sec:dataset}
We evaluate our method on all publicly available multiple-clustering benchmarks: Card \cite{yao2023augdmc}, CMUface \cite{gunnemann2014smvc}, Fruit \cite{hu2017finding}, Fruit360 \cite{yao2023augdmc}, and CIFAR10-MC \cite{yao2024customized}. The Card dataset contains 8,029 playing card images with two clustering criteria: order (Ace–King) and suits (clubs, diamonds, hearts, spades). CMUface includes 640 facial images annotated for pose (left, right, straight, up), identity (20 individuals), sunglasses (with/without), and emotion (angry, sad, happy, neutral). Fruit consists of 105 images clusterable by species (apple, banana, grape) or color (green, red, yellow). Fruit360 extends this with 4,856 images labeled by species (apple, banana, grape, cherry) and color (green, red, yellow, burgundy). CIFAR10-MC includes 60,000 images grouped by type (transportation, animals) and environment (land, air, water).

\subsection{Implementation Details}
\label{sec:implementation}
We adopt LLaVA \cite{liu2024visual} as our MLLM model, using Qwen2-7B \cite{yang2024qwen2technicalreport} as the language model. We fine-tune only the language model with LoRA \cite{hu2022lora}. Both the embedding extractor and the agent are trained for $50$ epochs on $4$ A100 GPUs over $3$ days, with a total batch size of $32$. Optimization is performed using AdamW \cite{loshchilov2017decoupled} with a learning rate of $1\mathrm{e}{-5}$ and a cosine decay schedule. We set the embedding similarity threshold $\tau$ to $0.6$ to filter weak edges. Clustering performance is evaluated using Normalized Mutual Information (NMI) \cite{white2004performance} and Rand Index (RI) \cite{rand1971objective}.

\subsection{Results}
As shown in Table~\ref{tab:1}, our method consistently outperforms state-of-the-art models across all multiple clustering benchmarks. On the Fruit dataset, it achieves perfect NMI and RI scores ($1.0000$) for both color- and species-based clustering. On Fruit360, it attains an NMI of $0.7214$ for color-based clustering, exceeding Multi-Map and Multi-Sub by $15.6\%$ and $8.4\%$, respectively, and an NMI of $0.6532$ for species-based clustering, outperforming them by $24.3\%$ and $6.7\%$. On CIFAR10-MC, our method achieves an NMI of $0.6385$ and RI of $0.7812$ for type-based clustering, surpassing Multi-Sub by $21.1\%$ and $5.7\%$, and an NMI of $0.5931$ and RI of $0.7439$ for environment-based clustering, exceeding Multi-Sub by $22.8\%$ and $4.8\%$. On Card, it reaches $0.9667$ NMI and $0.9952$ RI for order-based clustering, and $0.9481$ NMI and $0.9882$ RI for suits-based clustering, outperforming Multi-Sub by up to $205.4\%$ in NMI and $24.4\%$ in RI. On CMUface, it achieves $0.9720$ NMI and $0.9936$ RI for sunglass-based clustering, and $0.7789$ NMI and $0.9002$ RI for pose-based clustering, showing up to $99.6\%$ and $31.5\%$ improvements over Multi-Sub.

These significant improvements highlight the effectiveness of our agent-centric clustering framework in capturing user interests (e.g., order, suits, sunglass, pose) with advanced contextual understanding capabilities from MLLMs, while minimizing interference from irrelevant information.

\begin{figure}[t]
\centering
\includegraphics[bb=0 0 2091 627, width=1.0\linewidth]{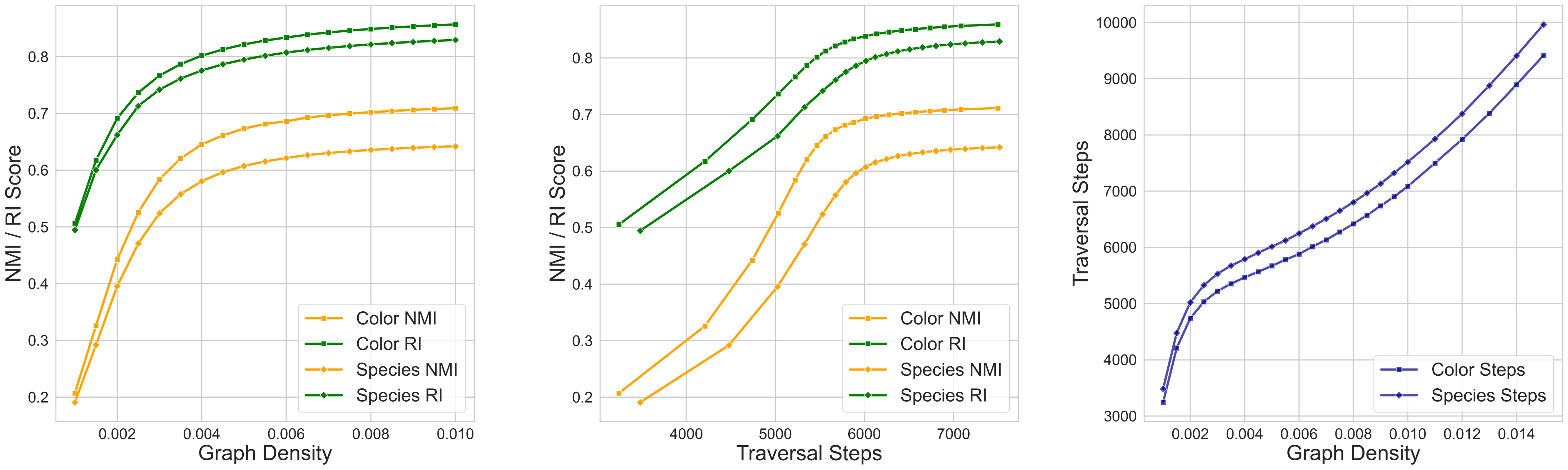}
\caption{Graph density vs. clustering metrics and traversal steps.}
\label{Fig:graph_density}
% \vspace{-4mm}
\end{figure}

\begin{figure}[t]
\centering
\includegraphics[bb=2 3 1015 212, width=1.0\linewidth]{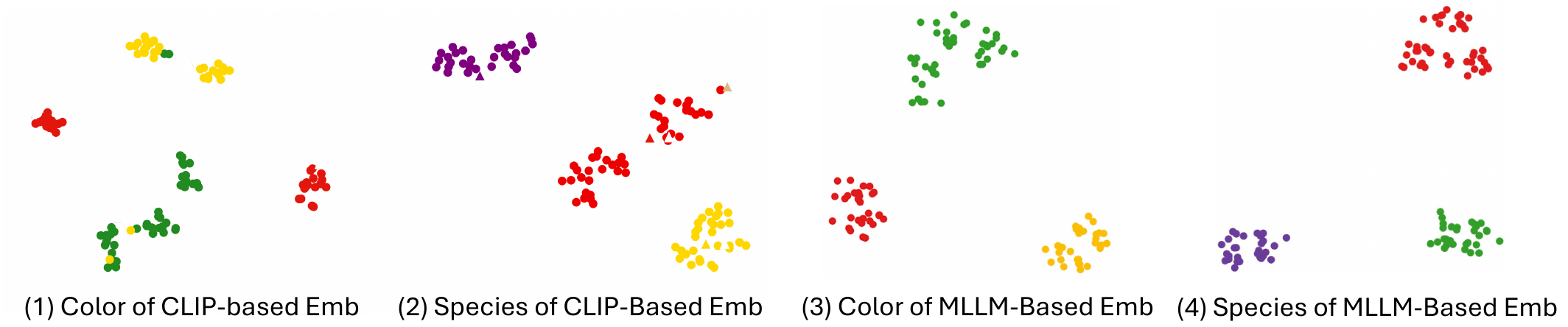}
\caption{T-SNE visualization of MLLM-based and CLIP-based embeddings on the Fruit dataset.}
% with corresponding labels on the Fruit dataset. For color-based embeddings, red, green, and yellow points correspond to fruits that are red, green, and yellow, respectively. For species-based embeddings, red, yellow, and purple points correspond to apples, bananas, and grapes, respectively.}
\label{Fig:embedding_visualization}
\vspace{-4mm}
\end{figure}

\subsection{Ablation Studies}

\paragraph{Density of Relational Graph.}
We investigate how graph density affects agent traversal by varying the embedding similarity threshold $\tau$ to construct relational graphs of different densities. Experiments are conducted on Fruit360. As shown in Fig.~\ref{Fig:graph_density}, increasing graph density (i.e., relaxing $\tau$) improves NMI and RI for both color- and species-based clustering until performance saturates. For instance, Color RI and Species RI quickly rise and plateau around $0.87$ and $0.84$, while Color NMI and Species NMI converge near $0.72$ and $0.65$. This suggests that once sufficient edges form connected components of semantically similar nodes, further densification yields diminishing returns.

However, denser graphs incur higher traversal costs. As graph density increases from $0.001$ to $0.015$, the number of agent traversal steps (dark blue curve) rises from $3,000$ to nearly $10,000$ due to a significant increase in spurious edges, indicating substantial computational overhead. A moderate $\tau$ strikes a balance: clustering accuracy stabilizes while reducing unnecessary traversal. These findings underscore the benefits of using MLLM-based embeddings to construct relational graphs. By pruning weak edges with low similarity scores, agents avoid exhaustive searches in irrelevant regions. At the same time, the context-aware nature of MLLM embeddings retains high-quality edges, aligning clusters with user interests while keeping traversal efficient.

\begin{table*}[t]
\caption{Ablation study on the Fruit360, Card, and CMUface datasets.}
\centering
\footnotesize  % 可根据需要改为 \scriptsize 等
\resizebox{0.9\textwidth}{!}{
\begin{tabular}{l l l c c c c c c c}
\toprule
\multirow{2}{*}{\textbf{Embedding}} & \multirow{2}{*}{\textbf{Dataset}} & \multirow{2}{*}{\textbf{Aspect}} & \multicolumn{2}{c}{\textbf{K-Means}} & \multicolumn{2}{c}{\textbf{HDBSCAN}} & \multicolumn{3}{c}{\textbf{Agent-Centric Graph Traversal}} \\
\cmidrule(lr){4-5}\cmidrule(lr){6-7}\cmidrule(lr){8-10}
 &  &  & NMI & RI & NMI & RI & NMI & RI & Traversal Steps \\
\toprule
\multirow{6}{*}{\textbf{CLIP Emb}}      & \multirow{2}{*}{Fruit360} & Color    & 0.6239 & 0.8243    & 0.6541 & 0.8408    & 0.6922 & 0.8573 & 14821 \\
                                        &                           & Species  & 0.5284 & 0.7582    & 0.5702 & 0.7851    & 0.6158 & 0.8180 & 14712 \\
\cmidrule(lr){2-10}
                                        & \multirow{2}{*}{Card}     & Order    & 0.3653 & 0.8587    & 0.4163 & 0.8708    & 0.8464 & 0.9679 & 29456 \\
                                        &                           & Suits    & 0.2734 & 0.7039    & 0.3368 & 0.7275    & 0.8132 & 0.9313 & 14969 \\
\cmidrule(lr){2-10}
                                        & \multirow{2}{*}{CMUface}  & Sunglass & 0.3402 & 0.7068    & 0.3952 & 0.7301    & 0.8456 & 0.9362 & 1281 \\
                                        &                           & Pose     & 0.4693 & 0.6624    & 0.4935 & 0.6804    & 0.7168 & 0.8526 & 4015 \\
\midrule
\multirow{6}{*}{\textbf{MLLM Emb}}      & \multirow{2}{*}{Fruit360} & Color    & 0.6629 & 0.8432    & 0.6912 & 0.8578    & \textbf{0.7214} & \textbf{0.8715} & \textbf{6404} \\
                                        &                           & Species  & 0.5783 & 0.7924    & 0.6138 & 0.8179    & \textbf{0.6532} & \textbf{0.8436} & \textbf{6964} \\
\cmidrule(lr){2-10}
                                        & \multirow{2}{*}{Card}     & Order    & 0.7262 & 0.9406    & 0.7743 & 0.9515    & \textbf{0.9667} & \textbf{0.9952} & \textbf{12837} \\
                                        &                           & Suits    & 0.6782 & 0.8745    & 0.7322 & 0.8972    & \textbf{0.9481} & \textbf{0.9882} & \textbf{9970} \\
\cmidrule(lr){2-10}
                                        & \multirow{2}{*}{CMUface}  & Sunglass & 0.7193 & 0.9005    & 0.7698 & 0.9191    & \textbf{0.9720} & \textbf{0.9936} & \textbf{798} \\
                                        &                           & Pose     & 0.6551 & 0.8051    & 0.6799 & 0.8241    & \textbf{0.7789} & \textbf{0.9002} & \textbf{1452} \\
\bottomrule
\end{tabular}
}
% Agent-centric graph traversal clearly outperforms K-Means and HDBSCAN on both kinds of embeddings. Meanwhile, MLLM embeddings show significant advantages over CLIP embedding in reducing the traversal steps. }
\label{tab:2}
% \vspace{-3mm}
\end{table*}
%  We compare the proposed MLLM-based embeddings with CLIP-based embeddings across different clustering methods and evaluate the proposed agent-centric graph traversal clustering against K-means for both embeddings. The evaluation metrics include Normalized Mutual Information (NMI) and Rand Index (RI). Additionally, we report the number of traversal steps for agent-centric graph traversal clustering.

\begin{table}[t]
\caption{Ablation study on number of representative nodes $K$.}
\centering
\small
\resizebox{0.9\linewidth}{!}{
\begin{tabular}{c |c c c c|c c c c}
\toprule
\multirow{2}{*}{\textbf{Num of Representatives}} & \multicolumn{2}{c}{Fruit360-Color} & \multicolumn{2}{c|}{Fruit360-Species} & \multicolumn{2}{c}{Card-Order} & \multicolumn{2}{c}{Card-Suits} \\
\cmidrule(lr){2-3}\cmidrule(lr){4-5}\cmidrule(lr){6-7}\cmidrule(lr){8-9}
& NMI & RI & NMI & RI & NMI & RI & NMI & RI \\
\toprule
$K=1$ & 0.7131 & 0.8643 & 0.6424 & 0.8361   & 0.9625 & 0.9930 & 0.9421 & 0.9860 \\
$K=2$ & 0.7185 & 0.8691 & 0.6490 & 0.8402   & 0.9652 & 0.9943 & 0.9461 & 0.9873 \\
$K=3$ & 0.7214 & 0.8715 & 0.6532 & 0.8436   & 0.9667 & 0.9952 & 0.9481 & 0.9882 \\
$K=4$ & 0.7240 & 0.8730 & 0.6559 & 0.8457   & 0.9673 & 0.9957 & 0.9492 & 0.9888 \\
\bottomrule
\end{tabular}
}
\label{tab:3}
% \vspace{-3mm}
\end{table}

\paragraph{MLLM‐based vs. CLIP-based Embeddings.} We compare our MLLM-based embeddings with CLIP-based ones across various clustering methods. We use Multi-Map \cite{yao2024multi} as the CLIP baseline, as both approaches decouple embedding extraction from clustering for fair ablation. As shown in Table~\ref{tab:2}, MLLM-based embeddings consistently outperform CLIP-based ones. For example, on CMUface (``Sunglass'' aspect), MLLM-based embeddings improve NMI from $0.3402$, $0.3952$, and $0.8456$ to $0.7193$, $0.7698$, and $0.9720$ for K-means, HDBSCAN, and agent-centric clustering, respectively—demonstrating superior alignment with user-defined criteria. Fig.~\ref{Fig:embedding_visualization} further illustrates this: MLLM-based embeddings yield compact, well-separated clusters in t-SNE visualization on the Fruit dataset, while CLIP-based embeddings show higher intra-cluster variance, particularly in color-based clustering. These results confirm that MLLM embeddings more effectively capture user interests (e.g., color, species), leading to improved clustering accuracy.

Moreover, MLLM-based embeddings significantly reduce agent traversal steps in agent-centric clustering. On Fruit360 (``Color'' aspect), agent traversal completes in $6,404$ steps with MLLM embeddings, compared to $14,821$ with CLIP-based ones. This efficiency arises from the ability of MLLM embeddings to better prune noisy edges, retaining a compact set of high-quality neighbors and reducing the number of membership assessments needed during graph traversal.

\paragraph{Agent-Centric Clustering vs. K-Means \& HDBSCAN.} We compare our agent-centric graph traversal clustering with classical K-Means and HDBSCAN. As shown in Table~\ref{tab:2}, the agent-centric approach consistently outperforms both baselines in NMI and RI across all datasets and embedding types. On the Card dataset (``Order'' aspect) with MLLM-based embeddings, it achieves an NMI of $0.9667$, outperforming K-Means ($0.7262$) and HDBSCAN ($0.7743$). The improvement is even larger with CLIP-based embeddings, reaching an NMI of $0.8464$ vs. $0.3653$ (K-Means) and $0.4163$ (HDBSCAN), demonstrating the robustness of the agent-centric approach. Furthermore, it requires fewer traversal steps with MLLM-based embeddings than with CLIP-based ones, improving efficiency.

\paragraph{Number of Representative Nodes.} Table~\ref{tab:3} shows the effect of varying $K$ on clustering performance (NMI and RI). Using a single representative node may fail to capture the cluster’s diversity, leading to lower performance. Increasing $K$ improves both metrics by better reflecting the cluster’s structure, although gains beyond three or four representatives are typically modest. Thus, a moderate $K$ strikes the best balance between capturing diversity and avoiding redundant information.

\section{Impacts and Limitations}
\label{Sec:limitations}
Our main limitation is relying on GPT-4 for generating pseudo labels for MLLM's embedding and agent training, as current open-source MLLMs like LLaVA lack the ability to produce accurate or semantically rich descriptions. Future work will explore stronger open-source MLLMs capable of generating their own supervision signals, enabling self-supervised embedding training. This paradigm offers a flexible and scalable solution for clustering. Our approach supports customizable clustering criteria and is well-suited for industrial applications such as search and graph mining.

\section{Conclusion}
This paper presents an agent-centric clustering framework that leverages MLLMs to traverse relational graphs and search for clusters based on user interests. By leveraging MLLMs' reasoning abilities, our method better captures user clustering preferences than CLIP-based approaches. To improve efficiency, we construct relational graphs from user-interest-biased embeddings derived from MLLMs, and reduce agents' traversal steps by filtering weak connections. Experiments on multiple benchmarks demonstrate the state-of-the-art performance of our method, validating its effectiveness.

\clearpage

\bibliographystyle{plain}
\bibliography{neurips_2025}

%%%%%%%%%%%%%%%%%%%%%%%%%%%%%%%%%%%%%%%%%%%%%%%%%%%%%%%%%%%%

\clearpage

\appendix

\section*{Appendix}

The Appendix is organized as follows:
\begin{itemize}
    % \item Appendix \ref{Sec:appendix_A}: More implementation details of our proposed framework, including key code snippets for multi-modal large language model, embedding generation, graph construction, agent traversal, and data processing.

    \item Appendix \ref{Sec:appendix_A}: Additional implementation details of our proposed framework, including key code snippets and explanations for each core component:
    
    \begin{itemize}
        \item \textbf{Embedding \& Agent Training Data Generation} (\ref{sec:data_genetation}): Describes how to generate labeled data for embedding and agent training with GPT-4.
        
        \item \textbf{Multi-Modal Large Language Model} (\ref{sec:model_definition}): Describes the initialization and LoRA-based fine-tuning of the MLLM used for both embedding extraction and agent-based decision making.
        
        \item \textbf{MLLM-Based Image Embedding Extraction} (\ref{sec:embedding_extraction}): Details the process of generating user-interest-biased image embeddings by extracting hidden states associated with the \texttt{<embedding>} token.
        
        \item \textbf{MLLM-Based Image Embedding Training} (\ref{sec:embedding_training}): Explains how image embeddings are fine-tuned using GPT-4-generated captions and similarity labels, supervised with cross-entropy and contrastive loss.
        
        \item \textbf{MLLM-Embedding-Based Graph Construction} (\ref{sec:mllm_graph_construction}): Outlines the construction of a sparse relational graph using MLLM embeddings and a similarity threshold, to guide agent traversal.
        
        \item \textbf{Agent-Based Node Membership \& Cluster Merge Assessment} (\ref{sec:agent_assessment}): Presents the use of MLLM-generated textual conclusions to assess candidate node membership and determine whether clusters should be merged.
        
        \item \textbf{Agent-Centric Graph Traversal} (\ref{sec:agent_graph_traversal}): Describes how agents iteratively expand clusters within each connected component based on MLLM-guided evaluations of neighboring nodes.
        
        \item \textbf{Agent-Centric Cluster Merge} (\ref{sec:agent_cluster_merge}): Details the global refinement stage in which semantically redundant clusters are merged through agent-based inter-cluster similarity assessments.
    \end{itemize}
    
    \item Appendix \ref{Sec:appendix_B}: Additional qualitative examples that highlight the interpretability and effectiveness of our agent-based clustering framework. 
    
    \begin{itemize}
        \item \textbf{Examples of Agent-Based Node Membership Assessment} (\ref{sec:agent_node_assessment_example}): Showcases how agents assess candidate nodes across different clustering aspects (e.g., number, suits, color) by reasoning over visual features and user-defined criteria. Incorrect connections in the relational graph are progressively pruned through these assessments, improving clustering purity.
    
        \item \textbf{Examples of Agent-Based Cluster Merge Assessment} (\ref{sec:agent_cluster_assessment_example}): Demonstrates how agents evaluate whether semantically similar clusters should be merged. The examples illustrate how missing edges in the graph are recovered via agent-guided assessments, enhancing the cohesion and completeness of the final clustering results.
    \end{itemize}
\end{itemize}

\section{More Framework Details with Key Code Snippets}
\label{Sec:appendix_A}

\subsection{Embedding \& Agent Training Data Generation}
\label{sec:data_genetation}
The embedding and agent training data are generated dynamically during the training process. For embedding training, in each epoch, we first extract user-interest-biased image embeddings using the MLLM, as described in Sec.\ref{sec:embedding_extraction}. We then perform hard negative mining by selecting the top $1,024$ image pairs with the highest embedding similarity entropy. Over $50$ training epochs, this results in approximately $50,000$ image pairs. As shown in the provided code snippet, GPT-4 is used to evaluate the semantic similarity of these image pairs based on user-defined criteria, from which binary pseudo labels are inferred. These labels supervise the pairwise embedding similarities using a binary cross-entropy loss, as detailed in Sec.\ref{sec:embedding_training}. In addition, GPT-4 generates detailed captions for each image, appending a special \texttt{<embedding>} token at the end. These captions serve as targets for training the MLLM's language generation via cross-entropy loss. The hidden state corresponding to the \texttt{<embedding>} token in the MLLM output is then projected to form the final image embedding, which captures user-interest-aligned semantics.

For agent training, we construct a relational graph in each epoch using embeddings extracted by the MLLM, as described in Sec.\ref{sec:mllm_graph_construction}. From this graph, we uniformly sample a subgraph containing $1,024$ nodes. GPT-4 then traverses this subgraph following the procedure in Sec.\ref{sec:agent_graph_traversal}, evaluating the membership of each neighboring node with respect to the current cluster based on user-defined interests. These membership assessments are collected at each traversal step and used as training samples. The membership assessment with GPT4 is similar to ``generate\_image\_similarity\_with\_GPT4'' in the following code snippet. During training, the MLLM is tasked with predicting whether each sampled neighboring node belongs to its associated cluster, conditioned on user interests, as described in Sec.~\ref{sec:agent_assessment}. The model's predictions are supervised using GPT-4's decisions, optimized via cross-entropy loss.

\begin{lstlisting}[caption={Embedding Training Data Generation Code Snippet}]
openai_api_key = "xxx"

prompt_caption_color = "Describe the color of the fruit in the provided image in detail."

prompt_similarity_color = "Determine whether the two fruits share the same color. Respond with <CONCLUSION> YES </CONCLUSION> or <CONCLUSION> NO </CONCLUSION>."

def encode_image_base64(path):
    with open(path, "rb") as image_file:
        return base64.b64encode(image_file.read()).decode("utf-8")

def generate_image_caption_with_GPT4(image_path):
    base64_image = encode_image_base64(image_path)
    client = openai.OpenAI(api_key=openai_api_key)
    
    response = client.chat.completions.create(
        model="gpt-4o",
        messages=[
            {
                "role": "user",
                "content": [
                    {"type": "text", "text": prompt_caption_color},
                    {
                        "type": "image_url",
                        "image_url": {
                            "url": f"data:image/png;base64,{base64_image}"
                        }
                    }
                ]
            }
        ]
    )
    caption = response.choices[0].message.content + "Embedding: <embedding>"
    return caption

def generate_image_similarity_with_GPT4(image_path1, image_path2)
    base64_image1 = encode_image_base64(image_path1)
    base64_image2 = encode_image_base64(image_path2)
    client = openai.OpenAI(api_key=openai_api_key)
    
    response = client.chat.completions.create(
        model="gpt-4o",
        messages=[
            {
                "role": "user",
                "content": [
                    {"type": "text", "text": prompt_similarity_color},
                    {
                        "type": "image_url",
                        "image_url": {
                            "url": f"data:image/png;base64,{base64_image1}",
                            "detail": "auto"
                        }
                    },
                    {
                        "type": "image_url",
                        "image_url": {
                            "url": f"data:image/jpeg;base64,{base64_image2}",
                            "detail": "auto"
                        }
                    }
                ]
            }
        ],
        max_tokens=500
    )
    img_similarity_reason = response.choices[0].message.content
    match = re.search(r"<CONCLUSION>(.*?)</CONCLUSION>", img_similarity_reason, re.DOTALL)
    if match:
        conclusion = match.group(1).strip()
        if "yes" in conclusion.lower():
            img_similarity_label = 1.0
        elif "no" in conclusion.lower():
            img_similarity_label = 0.0
        else:
            img_similarity_label = -1.0
            print(f"No answer found in {conclusion}.")
    else:
        img_similarity_label = -1.0
        print(f"No conclusion found in {output}.")
    return img_similarity_label


caption_dict = {}
similarity_dict = {}

def generate_embedding_training_data_with_GPT4(sample_pairs):
    targets = []
    for image_path1, image_path2 in sample_pairs:
        if image_path1 not in caption_dict:
            caption_dict[image_path1] = generate_image_caption_with_GPT4(image_path1)
        if image_path2 not in caption_dict:
            caption_dict[image_path2] = generate_image_caption_with_GPT4(image_path2)
        if (image_path1, image_path2) not in similarity_dict:
            similarity_dict[(image_path1, image_path2)] = generate_image_similarity_with_GPT4(image_path1, image_path2)

        target = {
            "caption1": caption_dict[image_path1],
            "caption2": caption_dict[image_path2],
            "img_similarity_label": similarity_dict[(image_path1, image_path2)],
        }
        targets.append(target)
    return targets
\end{lstlisting}

\subsection{Multi-Modal Large Language Model}
\label{sec:model_definition}
The Multi-Modal Large Language Model (MLLM) serves as a crucial component in the proposed agent-centric personalized multiple clustering framework. In this setup, the MLLM is used both as an embedding extractor and an agent, assisting in the traversal of relational graphs to search for clusters based on user-specific interests. As detailed in the provided code snippet, the model is initialized with a pre-trained version of LlavaQwenForCausalLM, incorporating a vision tower for multimodal capabilities, enabling it to generate embeddings that reflect both visual and textual data.

A distinctive feature of the approach is the fine-tuning of the language model using LoRA, which ensures that only specific layers, including the language model head and embedding layers, are optimized. This selective fine-tuning contributes to the model's efficiency by focusing on task-relevant parts of the network. Moreover, the inclusion of an embedding token (the specialized token \texttt{<embedding>}) enables the model to transform the generated detailed image descriptions, aligned with user-defined criteria, into embeddings that are biased toward user interests. These embeddings form the backbone of the relational graph, where nodes represent data instances, and edges encode the similarities between them.

\begin{lstlisting}[caption={Multi-Modal Large Language Model Code Snippet}]
import torch
import torch.nn as nn
import networkx as nx
from llava.model import LlavaQwenForCausalLM
from peft import LoraConfig, get_peft_model

class MultipleClusteringLlavaQwen(nn.Module):
    # Model definition
    def __init__(self):
        super(MultipleClusteringLlavaQwen, self).__init__()
        # MLLM initialization
        self.llm = LlavaQwenForCausalLM.from_pretrained(
            "LLaVA-Video-7B-Qwen2",
            attn_implementation="flash_attention_2",
            torch_dtype=torch.bfloat16,
            low_cpu_mem_usage=False,
        )
        self.llm.get_model().initialize_vision_modules()
        vision_tower = self.llm.get_vision_tower()
        vision_tower.to(dtype=torch.bfloat16)
        
        # LORA configuration
        lora_config = LoraConfig(
            r=64,
            lora_alpha=16,
            target_modules=self.find_all_linear_names(),
            lora_dropout=0.05,
            bias="none",
            task_type="CAUSAL_LM",
        )
        self.llm = get_peft_model(self.llm, lora_config)
        self.llm.base_model.model.lm_head.requires_grad_(True)
        self.llm.base_model.model.model.embed_tokens.requires_grad_(True)
        self.llm.base_model.model.model.mm_projector.requires_grad_(True)

        # Embedding token definition
        self.tokenizer = AutoTokenizer.from_pretrained(
            "LLaVA-Video-7B-Qwen2", 
            cache_dir=None, 
            model_max_length=32768, 
            padding_side="right",
            use_fast=True,
        )
        self.tokenizer.add_tokens(["<embedding>"], special_tokens=True)
        self.embedding_token_index = self.tokenizer.convert_tokens_to_ids("<embedding>")

        # Embedding projection layer
        self.ln_final = nn.LayerNorm(3584)
        self.emb_proj = nn.Parameter(torch.empty(3584, 768))
        nn.init.normal_(self.emb_proj, std=3584 ** -0.5)
        self.logit_scale = nn.Parameter(torch.ones([]) * np.log(1 / 0.07))

        # Embedding similarity threshold to filter weak connections
        self.embedding_similarity_threshold = 0.6
        # Number of reprentative nodes for each cluster
        self.num_centroids = 3
        # Number of candidate neighbouring nodes for each cluster
        self.num_candidates = 1
        
        # MLLM Embedding list
        self.embedding_list = []
        # Agent neighbouring node membership assessment dict
        self.agent_local_assessment_dict = {}
        # Agent neighbouring cluster merge assessment dict
        self.agent_global_assessment_dict = {}
        # Relational graph for agent traversal
        self.relational_graph = nx.Graph()

        # Connected components for the relational graph
        self.communities = []
        # Current cluster for each connected component
        self.local_clusters = []
        # Reprentative nodes for each current cluster
        self.local_centroids = []
        # Candidate neighboring nodes for each current cluster
        self.local_candidates = []

        # Searched clusters from different connected components
        self.global_clusters = []
        # Reprentative nodes for each searched cluster
        self.global_centroids = []
        # Candidate neighboring clusters for each searched cluster
        self.global_candidates = []
        
    # Find LORA layers
    def find_all_linear_names(self):
        cls = torch.nn.Linear
        lora_module_names = set()
        multimodal_keywords = ["mm_projector", "vision_tower", "vision_resampler"]
        for name, module in self.llm.named_modules():
            if any(mm_keyword in name for mm_keyword in multimodal_keywords):
                continue
            if isinstance(module, cls):
                lora_module_names.add(name)
        if "lm_head" in lora_module_names:  # needed for 16-bit
            lora_module_names.remove("lm_head")
        return list(lora_module_names)
\end{lstlisting}

\subsection{MLLM-Based Image Embedding Extraction}
\label{sec:embedding_extraction}
The MLLM-Based Image Embedding Extraction process is crucial for generating user-interest-biased image representations in the agent-centric personalized clustering framework. As detailed in the provided code snippet, the model takes input data, including images and user-defined criteria, and generates detailed image descriptions aligned with those interests. The hidden states corresponding to the \texttt{<embedding>} token in the generated descriptions serve as image embeddings, representing the images in a compact form that reflects user preferences.

The embeddings are further refined through layer normalization and projection to align them with user-defined criteria during training. They are then normalized to facilitate similarity calculation and relational graph construction, supporting efficient agent-centric graph traversal for personalized clustering. In contrast to CLIP-based embeddings, which are constrained by coarse-grained modality alignment, this approach leverages the MLLM's reasoning capabilities to generate contextually rich embeddings that focus on user-specified aspects and attend to relevant visual details, thereby improving clustering accuracy.

\begin{lstlisting}[caption={MLLM-Based Image Embedding Extraction Code Snippet}]
import torch
import torch.nn as nn

class MultipleClusteringLlavaQwen(nn.Module):
    # Extract user-interest-biased embeddings
    def extract_embeddings(self, data):
        result = self.llm.generate(
            inputs=data["input_ids"],
            images=data["images"],
            image_sizes=data["image_sizes"],
            modalities=data["modalities"],
            position_ids=None,
            attention_mask=data["attention_mask"],
            do_sample=True,
            temperature=0.2,
            top_p=None,
            num_beams=1,
            # no_repeat_ngram_size=3,
            max_new_tokens=2048,
            use_cache=True,
            return_dict_in_generate=True)
        
        output_ids = result["sequences"]
        
        img_embeddings = torch.cat([x[-1] for x in result["hidden_states"]], dim=1)
        img_embeddings = img_embeddings[:, -output_ids.shape[1]:]
        img_embeddings = self.ln_final(img_embeddings)
        img_embeddings = img_embeddings @ self.emb_proj
        img_embeddings = F.normalize(img_embeddings, p=2, dim=-1)

        logit_scale = self.logit_scale.exp().clamp(max=100).cpu().numpy()
        
        for i in range(len(output_ids)):
            embedding_index = torch.where(output_ids[i] == self.embedding_token_index)[0]
            if len(embedding_index) > 0:
                img_embedding = img_embeddings[i][embedding_index[0] + 1]
                img_embedding = img_embedding.detach().float().cpu().numpy()
            else:
                img_embedding = None
            
            self.embedding_list.append({
                "sample_id": data["sample_id"],
                "img_embedding": img_embedding,
                "logit_scale": logit_scale,
            })
\end{lstlisting}

\subsection{MLLM-Based Image Embedding Training}
\label{sec:embedding_training}
The MLLM-Based Image Embedding Training process is designed to fine-tune image embeddings in alignment with user interests, thereby enabling more accurate and efficient clustering. As outlined in the provided code snippet, this process involves training the Multi-Modal Large Language Model (MLLM) on image pairs, where each image is associated with a caption (a detailed description based on user interests) generated by GPT-4, and each pair is assigned a binary similarity pseudo label inferred from GPT-4's similarity assessment.

During training, the model takes each image and user-defined criteria as input, with GPT-4's image caption (appended with an \texttt{<embedding>} token) serving as the target for the MLLM to regress. This is supervised using cross-entropy loss. The hidden states corresponding to the \texttt{<embedding>} token in the target are extracted as image embeddings, acting as the user-interest-biased representation of the image. These embeddings are refined through layer normalization and projection, then further normalized to ensure they are on the correct scale for similarity calculations. The similarities between the two image embeddings are computed using cosine similarity, scaled by the learnable logit\_scale parameter, and supervised with GPT-4's pseudo labels via binary cross-entropy loss. This contrastive-style supervision ensures that the image embeddings are fine-tuned to reflect the user interests, improving their effectiveness in downstream clustering tasks.

\begin{lstlisting}[caption={MLLM-Based Image Embedding Training Code Snippet}]
import torch
import torch.nn as nn

class MultipleClusteringLlavaQwen(nn.Module):
    # Train user-interest-biased embeddings
    def train_embeddings(self, data):
        outputs1 = self.llm(
            input_ids=data["caption1"]["input_ids"],
            attention_mask=data["caption1"]["attention_mask"],
            labels=data["caption1"]["labels"],
            images=data["caption1"]["images"],
            modalities=data["caption1"]["modalities"],
            image_sizes=data["caption1"]["image_sizes"],
            output_hidden_states=True,
            return_dict=True)

        outputs2 = self.llm(
            input_ids=data["caption2"]["input_ids"],
            attention_mask=data["caption2"]["attention_mask"],
            labels=data["caption2"]["labels"],
            images=data["caption2"]["images"],
            modalities=data["caption2"]["modalities"],
            image_sizes=data["caption2"]["image_sizes"],
            output_hidden_states=True,
            return_dict=True)
        
        img_embeddings1 = self.ln_final(outputs1["hidden_states"][-1]) @ self.emb_proj
        img_embeddings2 = self.ln_final(outputs2["hidden_states"][-1]) @ self.emb_proj
        
        img_embeddings1 = F.normalize(img_embeddings1, p=2, dim=-1)
        img_embeddings2 = F.normalize(img_embeddings2, p=2, dim=-1)

        img_embeddings1 = img_embeddings1[torch.where(data["caption1"]["labels"] == self.embedding_token_index)]
        img_embeddings2 = img_embeddings2[torch.where(data["caption2"]["labels"] == self.embedding_token_index)]
        
        logit_scale = self.logit_scale.exp().clamp(max=100)
        logits = torch.einsum("bd,bd->b", img_embeddings1, img_embeddings2)
        logits = logit_scale * logits
        
        loss_similarity = F.binary_cross_entropy_with_logits(logits, data["img_similarity_label"])

        loss_dict = {
            "caption": outputs["loss"],
            "similarity": loss_similarity,
        }
        return loss_dict
\end{lstlisting}

\subsection{MLLM-Embedding-Based Graph Construction}
\label{sec:mllm_graph_construction}
To enable efficient agent traversal and clustering, we construct a relational graph in which nodes represent input images and edges encode the pairwise similarity between their corresponding embeddings extracted by the MLLM. As shown in the provided code snippet, we first extract all valid embeddings and compute their cosine similarities, scaled by the logit\_scale parameter to adjust the sharpness of similarity scores. A sigmoid function is then applied to map raw similarity values into the range $\left[0, 1\right]$. To ensure graph sparsity and remove noisy or semantically weak connections, we filter out edges whose similarity scores fall below a predefined threshold.

A relational graph is then constructed by adding each image (identified by its sample\_id) as a node. High-confidence edges—those with similarity scores exceeding the threshold—are added between node pairs, with edge weights corresponding to their computed similarity. This results in a sparse, high-quality relational graph that preserves meaningful semantic relationships while eliminating spurious connections. By retaining only a compact set of high-quality neighbors, the graph significantly reduces the number of traversal steps required by agents to evaluate them, thereby improving clustering efficiency.

\begin{lstlisting}[caption={MLLM-Embedding-Based Graph Construction Code Snippet}]
import torch
import torch.nn as nn
import numpy as np

class MultipleClusteringLlavaQwen(nn.Module):
    def build_relational_graph(self):
        # Compute the similarity matrix
        img_embeddings = [sample["img_embedding"] for sample in self.embedding_list]
        valid_embeddings = np.array([embedding for embedding in img_embeddings if embedding is not None])
        logit_scale = self.embedding_list[0]["logit_scale"]
        valid_similarity_matrix = valid_embeddings @ valid_embeddings.T
        valid_similarity_matrix = valid_similarity_matrix * logit_scale
        valid_similarity_matrix = 1 / (1 + np.exp(-valid_similarity_matrix))

        # Filter weak and invalid edges
        mask = np.array([embedding is not None for embedding in img_embeddings])
        similarity_matrix = np.zeros((len(img_embeddings), len(img_embeddings)))
        similarity_matrix[mask[:, None] & mask[None, :]] = valid_similarity_matrix.ravel()
        similarity_matrix = np.triu(similarity_matrix, k=1)
        high_similarity_indices = np.argwhere(similarity_matrix >= self.embedding_similarity_threshold)
        
        # Construct the relational graph
        for sample in self.embedding_list:
            self.relational_graph.add_node(sample["sample_id"])
        for i, j in high_similarity_indices:
            sample_id1 = self.embedding_list[i]["sample_id"]
            sample_id2 = self.embedding_list[j]["sample_id"]
            similarity_score = similarity_matrix[i, j]
            self.relational_graph.add_edge(sample_id1, sample_id2, weight=similarity_score)
\end{lstlisting}

\subsection{Agent-Based Node Membership \& Cluster Merge Assessment}
\label{sec:agent_assessment}
To ensure that clustering decisions reflect fine-grained, user-specific semantics, we incorporate an agent-based assessment mechanism during both node membership evaluation and cluster merging. This process leverages the reasoning capabilities of the MLLM to guide decisions based on visual content and user-defined criteria.

For node membership assessment, each candidate node and its associated cluster centroids are encoded as visual inputs and passed to the MLLM, along with a prompt designed to elicit a binary decision. The model generates a textual response containing a structured \texttt{<CONCLUSION>} tag indicating whether the candidate should be included in the cluster. A response of ``yes'' results in a positive assignment (similarity = $1.0$), while ``no'' leads to rejection (similarity = $0.0$). Unrecognized or missing conclusions (similarity = $-1$) are ignored during agent graph traversal.

Similarly, cluster merge assessment involves presenting two sets of cluster centroids as inputs. The MLLM evaluates whether the two clusters represent redundant or semantically similar concepts. The decision, extracted from the \texttt{<CONCLUSION>} tag, determines whether the clusters should be merged or remain separate. These assessments are stored in dictionaries that guide subsequent relational graph updates and cluster refinement during the agent traversal process.

This agent-based assessment framework introduces semantic interpretability and user-interest alignment into the clustering pipeline, enabling more accurate and personalized outcomes.

\begin{lstlisting}[caption={Agent-Based Node Membership \& Cluster Merge Assessment Code Snippet}]
import torch
import torch.nn as nn
import numpy as np

class MultipleClusteringLlavaQwen(nn.Module):
    # Assess candidate node's membership in the local cluster
    def assess_node_membership_with_agents(self, data):
        result = self.llm.generate(
            inputs=data["input_ids"],
            # cluster representative images + candidate image
            images=data["images"],
            image_sizes=data["image_sizes"],
            modalities=data["modalities"],
            position_ids=None,
            attention_mask=data["attention_mask"],
            do_sample=True,
            temperature=0.2,
            top_p=None,
            num_beams=1,
            # no_repeat_ngram_size=3,
            max_new_tokens=2048,
            use_cache=True,
            return_dict_in_generate=True)
        
        output_ids = result["sequences"]
        stop_str = "<|im_end|>"
        outputs = self.tokenizer.batch_decode(output_ids, skip_special_tokens=True)
        for i, output in enumerate(outputs):
            output = output.strip()
            if output.endswith(stop_str):
                output = output[:-len(stop_str)]
            output = output.strip()

            match = re.search(r"<CONCLUSION>(.*?)</CONCLUSION>", output, re.DOTALL)
            if match:
                conclusion = match.group(1).strip()
                if "yes" in conclusion.lower():
                    img_similarity = 1.0
                elif "no" in conclusion.lower():
                    img_similarity = 0.0
                else:
                    img_similarity = -1.0
                    print(f"No answer found in {conclusion}.")
            else:
                img_similarity = -1.0
                print(f"No conclusion found in {output}.")
        
            self.agent_local_assessment_dict.update({
                # sample_pair: (local_centroids, candidate_node)
                data["sample_pair"]: img_similarity,
            })

    # Assess whether the two clusters should be merged
    def assess_cluster_merge_with_agents(self, data):
        result = self.llm.generate(
            inputs=data["input_ids"],
            # cluster1 + cluster2 representative images
            images=data["images"],
            image_sizes=data["image_sizes"],
            modalities=data["modalities"],
            position_ids=None,
            attention_mask=data["attention_mask"],
            do_sample=True,
            temperature=0.2,
            top_p=None,
            num_beams=1,
            # no_repeat_ngram_size=3,
            max_new_tokens=2048,
            use_cache=True,
            return_dict_in_generate=True)
        
        output_ids = result["sequences"]
        stop_str = "<|im_end|>"
        outputs = self.tokenizer.batch_decode(output_ids, skip_special_tokens=True)
        for i, output in enumerate(outputs):
            output = output.strip()
            if output.endswith(stop_str):
                output = output[:-len(stop_str)]
            output = output.strip()

            match = re.search(r"<CONCLUSION>(.*?)</CONCLUSION>", output, re.DOTALL)
            if match:
                conclusion = match.group(1).strip()
                if "yes" in conclusion.lower():
                    img_similarity = 1.0
                elif "no" in conclusion.lower():
                    img_similarity = 0.0
                else:
                    img_similarity = -1.0
                    print(f"No answer found in {conclusion}.")
            else:
                img_similarity = -1.0
                print(f"No conclusion found in {output}.")
        
            self.agent_global_assessment_dict.update({
                # sample_pair: (global_centroids, candidate_centroids)
                data["sample_pair"]: img_similarity,
            })
\end{lstlisting}

\subsection{Agent-Centric Graph Traversal}
\label{sec:agent_graph_traversal}
Once the relational graph is constructed, we employ agents to traverse the graph to discover clusters aligned with user interests. As illustrated in the following code snippet, each connected component of the graph is treated as an independent subgraph. For each component, we assign an agent to it and initialize a cluster by selecting the node with the highest degree (weighted by edge similarity), which serves as the initial centroid (i.e., representative node). The agents then iteratively expand their respective clusters by evaluating the membership of neighboring nodes.

Candidate neighbors are selected from the neighborhood of the current cluster within its component. An MLLM-based agent evaluates each candidate node against the cluster’s centroids based on user-defined criteria to determine whether it should be merged into the cluster. Positive candidates are added to the cluster, and the centroids of the cluster are updated by selecting the top-degree nodes. In contrast, negative candidates lead to the removal of their connecting edges to the cluster, reducing noise and potential interference in future traversal.

This iterative process continues until no more valid neighbors remain. If a cluster is completed, it is pushed to the global list, and the agent initializes a new cluster within the remaining subgraph. This method ensures that the traversal is both efficient and semantically meaningful, as decisions are guided by MLLM-based assessments. The agent-centric traversal strategy enables the discovery of fine-grained, user-interest-aligned clusters while dynamically refining the graph structure to suppress noisy or irrelevant connections.

\begin{lstlisting}[caption={Agent-Centric Graph Traversal Code Snippet}]
import torch
import torch.nn as nn
import networkx as nx
import numpy as np

class MultipleClusteringLlavaQwen(nn.Module):
    # Traverse the graph with agents to search for clusters
    def traverse_graph_with_agents(self):
        # Construct graph and initialize clusters
        if len(self.embedding_list) > 0:
            # Construct relational graph
            self.build_relational_graph()
            # Find connected components of the graph
            self.communities = list(nx.connected_components(self.relational_graph))
            self.communities = [community for community in self.communities if len(community) > 0]
            # Initialize a cluster for each component
            self.local_clusters = [[
                max(self.relational_graph.subgraph(community).degree(weight="weight"), key=lambda x: x[1])[0]
                ] for community in self.communities]
            # Initialize representative nodes for each cluster
            self.local_centroids = [tuple(cluster) for cluster in self.local_clusters]
            # Reset MLLM embedding list
            self.embedding_list = []
        
        # Update local clusters and relational graph
        if len(self.agent_local_assessment_dict) > 0:
            for i, (local_cluster, local_centroids, local_candidates) in enumerate(zip(self.local_clusters, self.local_centroids, self.local_candidates)):
                if local_candidates is None:
                    continue
                
                pos_candidates = []
                neg_candidates = []
                for candidate in local_candidates:
                    if self.agent_local_assessment_dict[(local_centroids, candidate)]:
                        pos_candidates.append(candidate)
                    else:
                        neg_candidates.append(candidate)

                if len(pos_candidates) > 0:
                    # update local cluster
                    local_cluster.extend(pos_candidates)
                    self.local_clusters[i] = local_cluster
                    # update local centroids
                    num_centroids = min(self.num_centroids, len(local_cluster))
                    subgraph = self.relational_graph.subgraph(local_cluster)
                    local_centroids = tuple([node for node, degree in sorted(subgraph.degree(weight="weight"), key=lambda x: x[1], reverse=True)[:num_centroids]])
                    self.local_centroids[i] = local_centroids

                if len(neg_candidates) > 0:
                    # update relational graph
                    edges_to_remove = list(nx.edge_boundary(self.relational_graph, local_cluster, neg_candidates))
                    self.relational_graph.remove_edges_from(edges_to_remove)
        
            # Reset node membership assessment dict
            self.agent_local_assessment_dict = {}

        # Find neighbors and select candidate for each local cluster
        self.local_candidates = []
        for i, (local_cluster, local_centroids, community) in enumerate(zip(self.local_clusters, self.local_centroids, self.communities)):
            # find local neighbors
            local_neighbors = self.find_local_cluster_neighbors(local_cluster, community)
            while len(local_neighbors) == 0 and len(community) != 0:
                # push local cluster to global clusters
                self.global_clusters.append(local_cluster)
                self.global_centroids.append(local_centroids)
                # remove local cluster from communities
                community = community - set(local_cluster)
                if len(community) > 0:
                    subgraph = self.relational_graph.subgraph(community)
                    local_cluster = [max(subgraph.degree(weight="weight"), key=lambda x: x[1])[0]]
                    local_centroids = tuple(local_cluster)
                    local_neighbors = self.find_local_cluster_neighbors(local_cluster, community)
                else:
                    local_cluster = local_neighbors = []
                    local_centroids = tuple(local_cluster)
            self.local_clusters[i] = local_cluster
            self.local_centroids[i] = local_centroids
            self.communities[i] = community
            self.local_candidates.append(local_neighbors if len(local_neighbors) > 0 else None)

        # Generate sample pairs for node membership assessment
        sample_pairs = []
        for i, (local_centroids, local_candidates) in enumerate(zip(self.local_centroids, self.local_candidates)):
            if local_candidates is not None:
                for candidate in local_candidates:
                    sample_pairs.append((local_centroids, candidate))
        return sample_pairs
    
    # Find neighbors of the local cluster
    def find_local_cluster_neighbors(self, local_cluster, community):
        local_cluster = set(local_cluster)
        subgraph = self.relational_graph.subgraph(community)
        neighbors = nx.node_boundary(subgraph, local_cluster)

        if len(neighbors) > self.num_candidates:
            subgraph = self.relational_graph.subgraph(local_cluster | neighbors)
            neighbors = [node for node, degree in sorted(subgraph.degree(weight="weight"), key=lambda x: x[1], reverse=True) if node in neighbors]
            neighbors = neighbors[:self.num_candidates]
        return neighbors
\end{lstlisting}

\subsection{Agent-Centric Cluster Merge}
\label{sec:agent_cluster_merge}
After local clusters are formed through agent-centric graph traversal with each connected component, we perform a global refinement step to merge semantically redundant clusters. As shown in the provided code snippet, each cluster is represented by a set of centroids (i.e., representative nodes), and the model identifies neighboring clusters as merge candidates by evaluating inter-cluster connectivity. For each candidate cluster pair, an MLLM-based agent assesses whether the two clusters should be merged based on user-defined criteria. If deemed redundant, the clusters are merged; otherwise, connecting edges are removed to reinforce separation.

To maintain index consistency, cluster pairs selected for merging are processed in reverse order of their indices. Once merged, the new cluster's centroids are updated by selecting the top-degree nodes within the merged subgraph. This merging process reduces redundancy and corrects connectivity errors that may have occurred during local graph traversal. By leveraging the MLLM's reasoning capabilities in the global stage, the framework produces clusters that are both well-separated and structurally coherent, while remaining closely aligned with fine-grained user interests.

\begin{lstlisting}[caption={Agent-Centric Cluster Merge Code Snippet}]
import torch
import torch.nn as nn
import networkx as nx
import numpy as np

class MultipleClusteringLlavaQwen(nn.Module):
    # Merge semantically redundant clusters with agents
    def merge_clusters_with_agents(self):
        # Update global clusters and relational graph
        if len(self.agent_global_assessment_dict) > 0:
            # Identify indices of cluster pairs to be merged
            to_merge = set()
            for i, (global_cluster, global_centroids, global_candidates) in enumerate(zip(self.global_clusters, self.global_centroids, self.global_candidates)):
                if global_candidates is None:
                    continue
                candidate_index = self.global_centroids.index(global_candidates)
                if candidate_index == i:
                    continue  # Avoid self-merging

                if self.agent_global_assessment_dict[(global_centroids, global_candidates)]:
                    to_merge.add(tuple(sorted((i, candidate_index))))
                else:
                    edges_to_remove = list(nx.edge_boundary(self.relational_graph, global_cluster, self.global_clusters[candidate_index]))
                    self.relational_graph.remove_edges_from(edges_to_remove)
            
            # Delete in reverse order to avoid index shifting
            removed = set()
            for i, candidate_index in sorted(to_merge, key=lambda x: x[1], reverse=True):
                if i in removed or candidate_index in removed:
                    continue
    
                merged_cluster = list(set(self.global_clusters[i]) | set(self.global_clusters[candidate_index]))
                self.global_clusters[i] = merged_cluster
    
                num_centroids = min(self.num_centroids, len(merged_cluster))
                subgraph = self.relational_graph.subgraph(merged_cluster)
                self.global_centroids[i] = tuple(
                    node for node, degree in sorted(subgraph.degree(weight="weight"), key=lambda x: x[1], reverse=True)[:num_centroids])
    
                # Delete the larger index c
                self.global_clusters.pop(candidate_index)
                self.global_centroids.pop(candidate_index)
                self.global_candidates.pop(candidate_index)
                removed.add(candidate_index)  # Record deleted index

            # Reset cluster merge assessment dict
            self.agent_global_assessment_dict = {}

        # Find neighbors and select candidate for each global cluster
        self.global_candidates = []
        for i, (global_cluster, global_centroids) in enumerate(zip(self.global_clusters, self.global_centroids)):
            # find global neighbors
            global_neighbors = self.find_global_cluster_neighbors(global_cluster, global_centroids)
            self.global_candidates.append(global_neighbors if len(global_neighbors) > 0 else None)

        # Generate sample pairs for cluster merge assessment
        sample_pairs = []
        for i, (global_centroids, global_candidates) in enumerate(zip(self.global_centroids, self.global_candidates)):
            if global_candidates is not None:
                sample_pairs.append((global_centroids, global_candidates))
        return sample_pairs
    
    # Find candidates of the global cluster to merge
    def find_global_cluster_neighbors(self, global_cluster, global_centroids):
        neighbors = []
        degrees = []
        for centroids, cluster in zip(self.global_centroids, self.global_clusters):
            if centroids == global_centroids:
                continue
            degree = sum(self.relational_graph.edges[u, v]["weight"] for u, v in nx.edge_boundary(self.relational_graph, global_cluster, cluster))
            neighbors.append(centroids)
            degrees.append(degree)
        if len(neighbors) > 0:
            neighbors = [neighbor for neighbor, degree in sorted(zip(neighbors, degrees), key=lambda x: x[1], reverse=True)]
            neighbors = neighbors[0]
        return neighbors
\end{lstlisting}

\section{Additional Examples of Model Inputs and Outputs}
\label{Sec:appendix_B}

\subsection{Examples of Agent-Based Node Membership Assessment}
\label{sec:agent_node_assessment_example}

Table~\ref{tab:appendix1} presents additional qualitative examples illustrating the interpretability of our agent-based node membership assessment across different clustering aspects, including number, suits, and color. Each example shows a set of cluster centroids and a candidate node, along with the corresponding prompt and MLLM-generated output. The agent evaluates the candidate's membership by reasoning over visual cues relevant to the specified aspect. For instance, in the ``Number'' and ``Suits'' aspects of playing cards, the agent focuses on rank or suit similarities, respectively, while ignoring irrelevant attributes. In the ``Color'' aspect of fruit images, the agent attends to subtle hue variations. These examples demonstrate the MLLM's ability to generate interpretable decisions aligned with user-defined criteria, validating the semantic alignment and explainability of our clustering process.

It is worth noting that incorrect edges in the relational graph are corrected through agents' graph traversal, as negative assessments lead to edge removals, progressively refining the graph structure and improving clustering purity.

\subsection{Examples of Agent-Based Cluster Merge Assessment}
\label{sec:agent_cluster_assessment_example}

Table~\ref{tab:appendix2} presents qualitative examples of agent-based cluster merge assessment across different clustering aspects, including number, suits, and color. Each case includes a set of cluster centroids, candidate centroids, a task-specific prompt, and the corresponding MLLM-generated decision. The agent evaluates whether the two clusters should be merged based on user-defined clustering preference. For instance, in the ``Number'' and ``Suits'' aspects, the agent compares rank or suit consistency across clusters, while in the ``Color'' aspect, it focuses on visual color alignment. The structured output includes a human-interpretable explanation and a final decision marked with a \texttt{<CONCLUSION>} tag. These examples demonstrate the model's ability to make semantically meaningful merge decisions and reinforce the interpretability of our agent-based framework.

It is worth noting that missing edges in the relational graph are corrected through agents' cluster merging, as positive assessments trigger the integration of semantically redundant clusters, enhancing the cohesion and completeness of the clustering structure.

\begin{table}[t]
\caption{Examples of Agent-Based Node Membership Assessment.}
\centering
\small
\renewcommand\arraystretch{1.6}
\resizebox{1.0\linewidth}{!}{
\begin{tabular}{|>{\centering\arraybackslash}m{1cm}|m{4cm}|>{\centering\arraybackslash}m{4.5cm}|>{\centering\arraybackslash}m{1.5cm}|m{4cm}|}
\hline
\textbf{Aspect} & \textbf{Prompt} & \textbf{Cluster Centroids} & \textbf{Candidate} & \textbf{MLLM Generation} \\
\hline
\multirow{3}{*}[-20mm]{\centering Number} 
& \multirow{3}{=}[-10mm]{\justifying Determine whether the candidate playing card should be included in the existing playing card cluster based on \textcolor{red}{\textbf{rank}}. Ignore the suits and focus only on rank comparison. Respond with \texttt{<CONCLUSION> YES </CONCLUSION>} or \texttt{<CONCLUSION> NO </CONCLUSION>}.}
& \imgcenter[4.5cm]{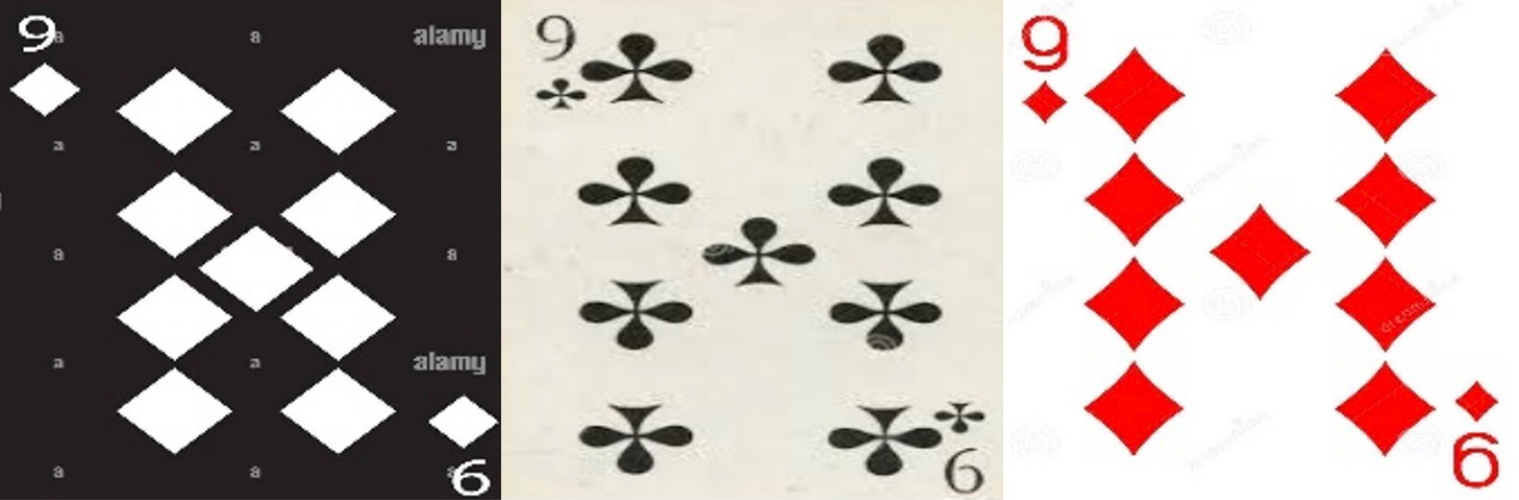}{0 0 729 240}
& \imgcenter[1.5cm]{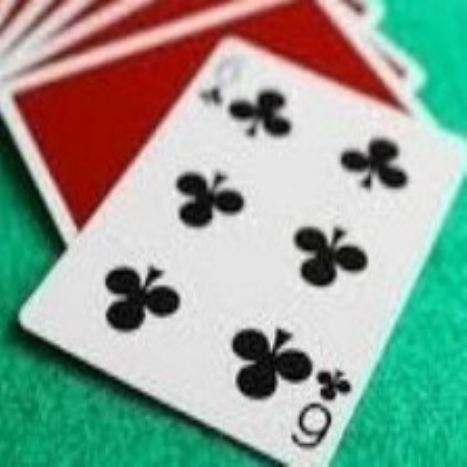}{0 0 224 224}
& The current cluster consists of cards with the rank 9. The candidate card is ranked 6. Based on rank difference, it should not be included. The answer is \texttt{<CONCLUSION>} \textcolor{red}{\textbf{NO}} \texttt{</CONCLUSION>}.\\

\cline{3-5}
& 
& \imgcenter[4.5cm]{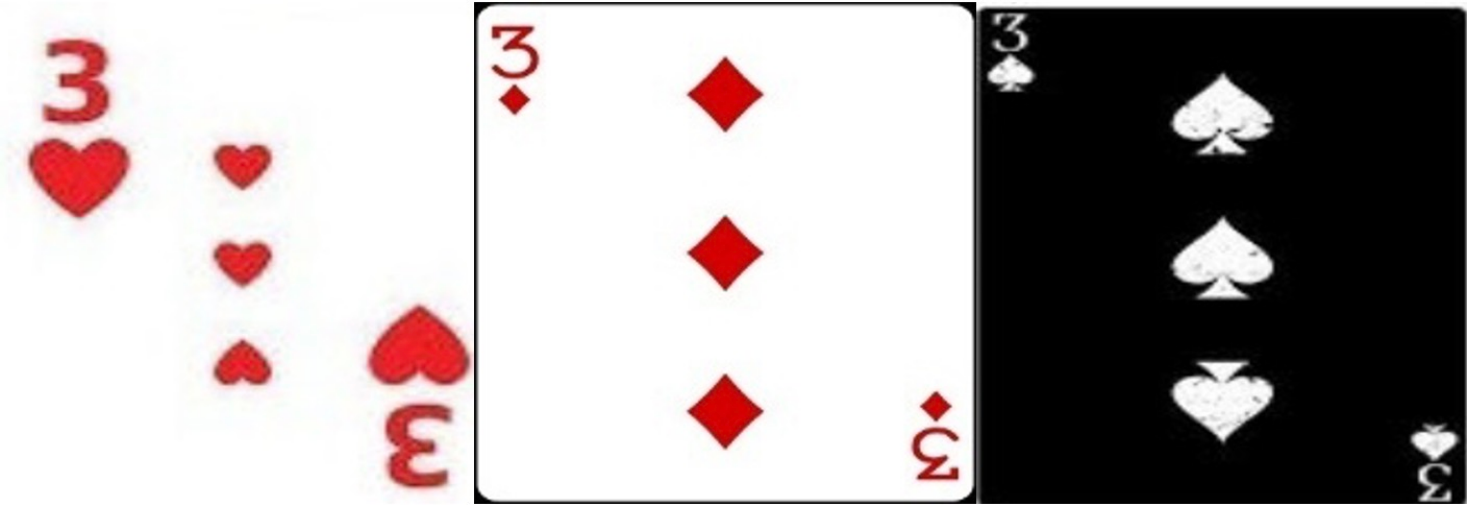}{0 0 704 242}
& \imgcenter[1.5cm]{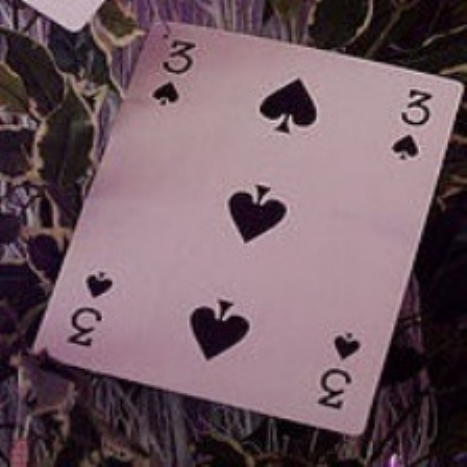}{0 0 224 224}
& The cluster contains cards of rank 3. The candidate card is also of rank 3. Based on rank similarity, it should be included. The answer is \texttt{<CONCLUSION>} \textcolor{red}{\textbf{YES}} \texttt{</CONCLUSION>}. \\

\cline{3-5}
& 
& \imgcenter[4.5cm]{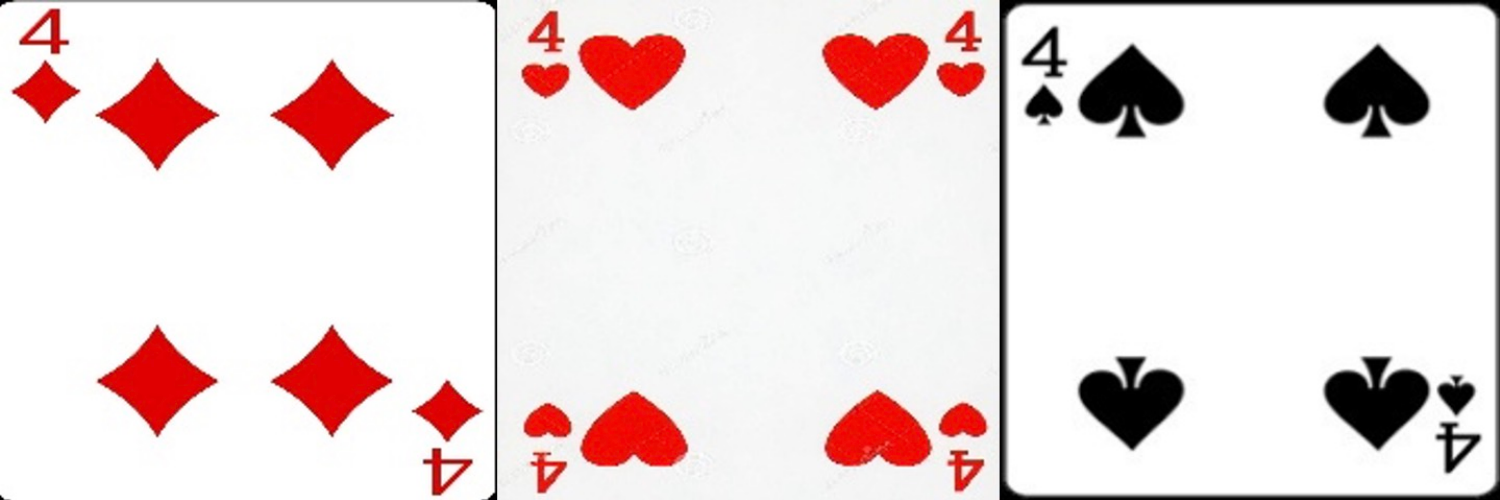}{0 0 720 240}
& \imgcenter[1.5cm]{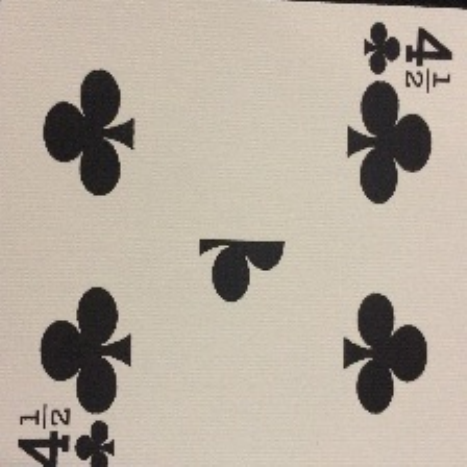}{0 0 224 224}
& The cluster contains cards of rank 4. The candidate card is also of rank 4. Based on rank similarity, it should be included. The answer is \texttt{<CONCLUSION>} \textcolor{red}{\textbf{YES}} \texttt{</CONCLUSION>}.\\
\hline

\multirow{3}{*}[-20mm]{\centering Suits} 
& \multirow{3}{=}[-10mm]{\justifying Determine whether the candidate playing card should be included in the existing playing card cluster based on \textcolor{red}{\textbf{suits}}. Ignore the rank and focus only on suits comparison. Respond with \texttt{<CONCLUSION> YES </CONCLUSION>} or \texttt{<CONCLUSION> NO </CONCLUSION>}.}
& \imgcenter[4.5cm]{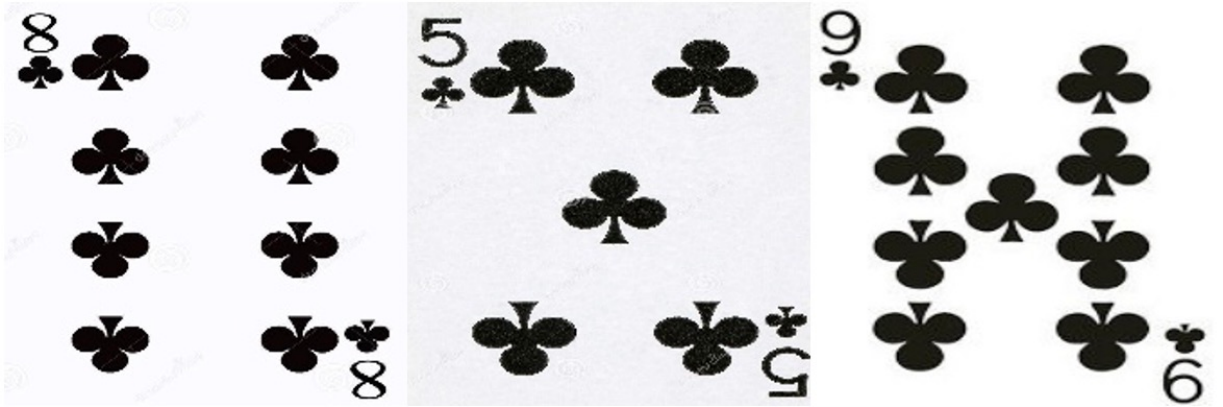}{0 0 587 195}
& \imgcenter[1.5cm]{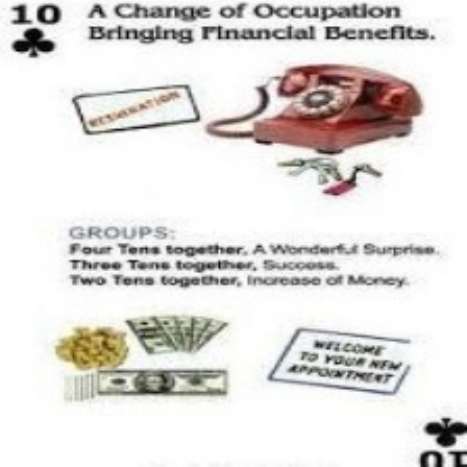}{0 0 224 224}
& The cluster contains cards of the club suit. The candidate card is also a club. Based on suit similarity, it should be included. The answer is \texttt{<CONCLUSION>} \textcolor{red}{\textbf{YES}} \texttt{</CONCLUSION>}.\\

\cline{3-5}
& 
& \imgcenter[4.5cm]{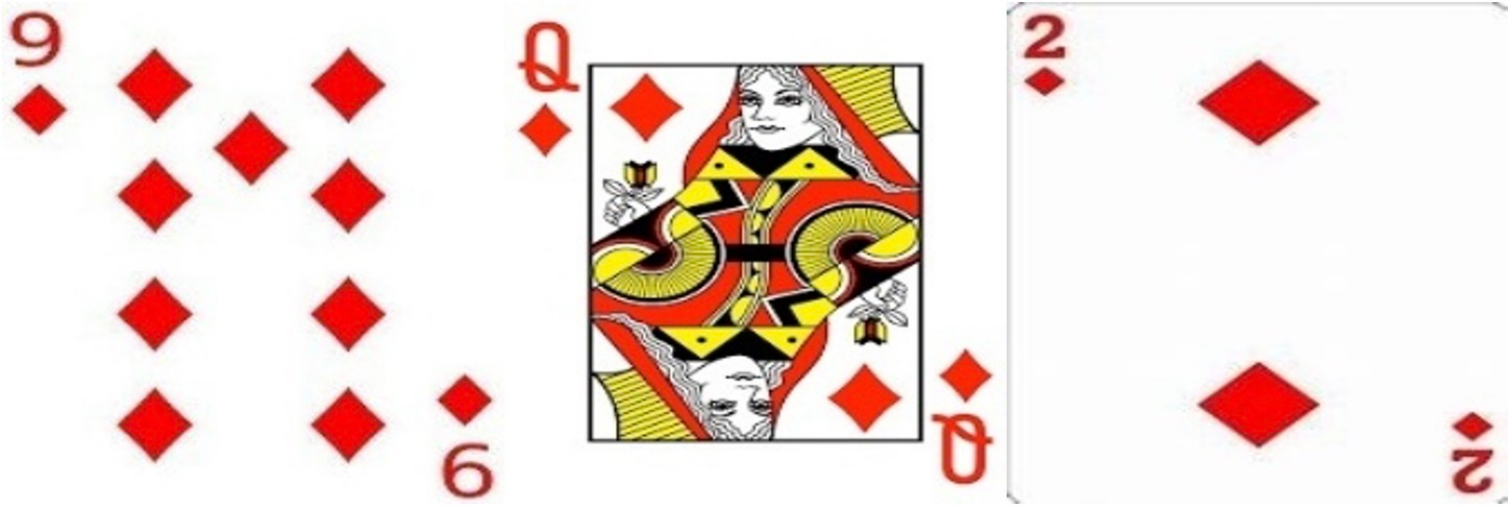}{0 0 724 243}
& \imgcenter[1.5cm]{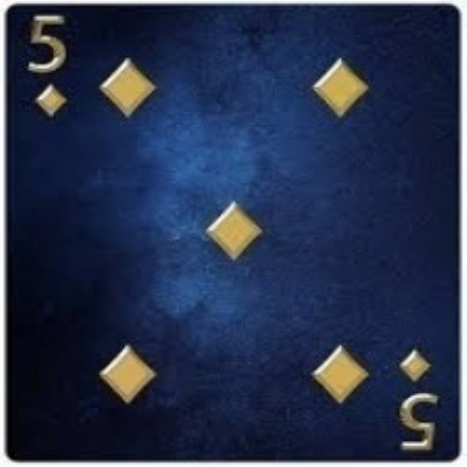}{0 0 224 224}
&  The cluster contains cards of the diamond suit. The candidate card is also a diamond. Based on suit similarity, it should be included. The answer is \texttt{<CONCLUSION>} \textcolor{red}{\textbf{YES}} \texttt{</CONCLUSION>}.\\

\cline{3-5}
& 
& \imgcenter[4.5cm]{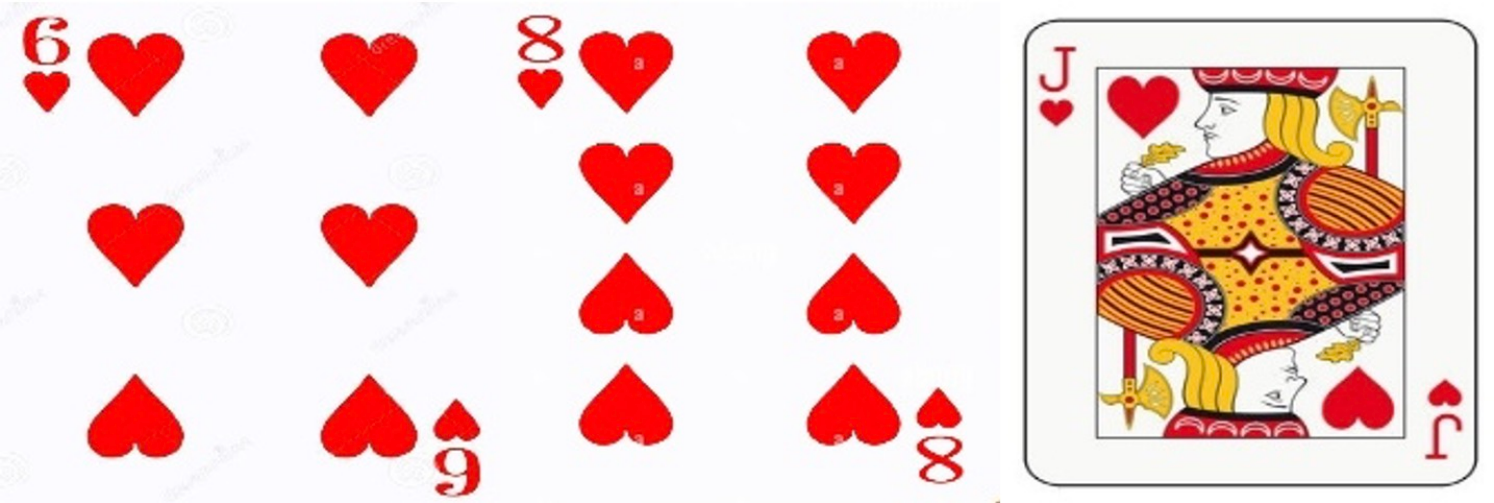}{0 0 717 242}
& \imgcenter[1.5cm]{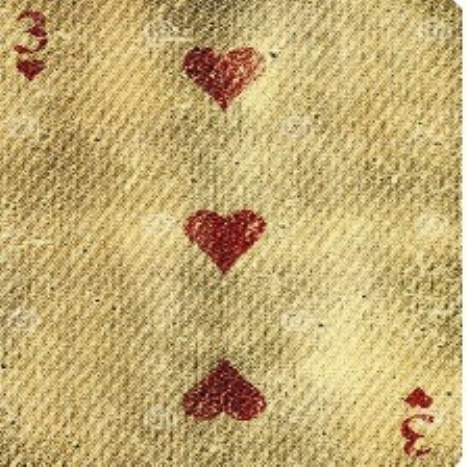}{0 0 224 224}
&  The cluster contains cards of the heart suit. The candidate card is also a heart. Based on suit similarity, it should be included. The answer is \texttt{<CONCLUSION>} \textcolor{red}{\textbf{YES}} \texttt{</CONCLUSION>}.\\
\hline

\multirow{3}{*}[-24mm]{\centering Color} 
& \multirow{3}{=}[-14mm]{\justifying Determine whether the candidate fruit should be included in the existing fruit cluster based on \textcolor{red}{\textbf{color}}
similarity. Ignore the species and focus only on color comparison. Respond with \texttt{<CONCLUSION> YES </CONCLUSION>}, or \texttt{<CONCLUSION> NO </CONCLUSION>}.}
& \imgcenter[4.5cm]{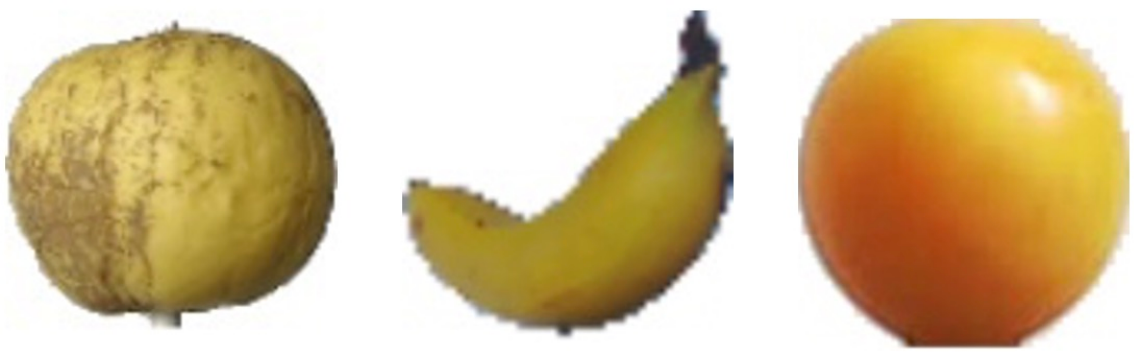}{0 0 548 170}
& \imgcenter[1.5cm]{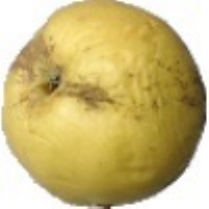}{0 0 101 101}
& The cluster consists of yellow-colored fruits. The candidate fruit also appears yellow. Based on color similarity, it should be included.
The answer is \texttt{<CONCLUSION>} \textcolor{red}{\textbf{YES}} \texttt{</CONCLUSION>}.\\

\cline{3-5}
& 
& \imgcenter[4.5cm]{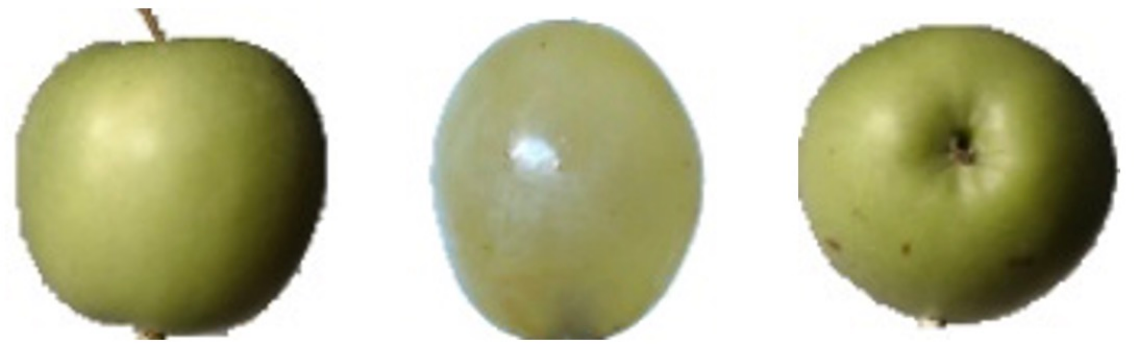}{0 0 546 166}
& \imgcenter[1.5cm]{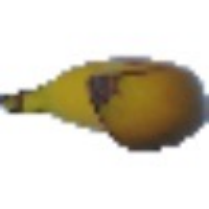}{0 0 101 101}
&  The cluster consists of green-colored fruits. The candidate fruit is yellow-green, which differs from the cluster’s dominant green tone. Based on color difference, it should not be included. The answer is \texttt{<CONCLUSION>} \textcolor{red}{\textbf{NO}} \texttt{</CONCLUSION>}.\\

\cline{3-5}
& 
& \imgcenter[4.5cm]{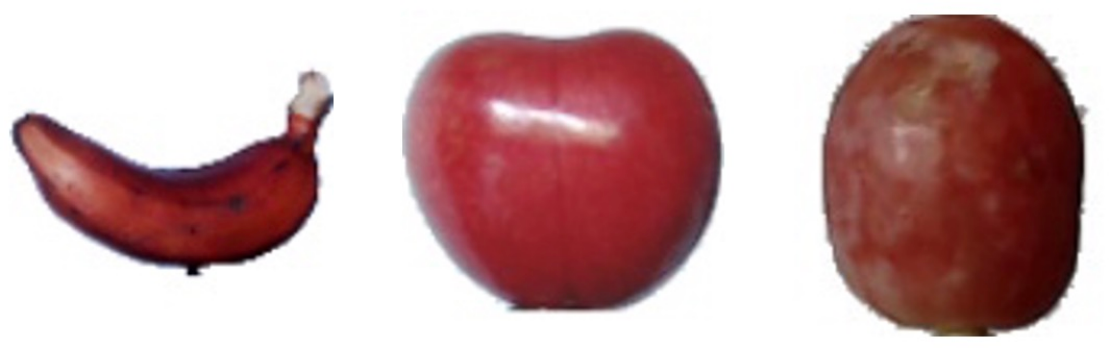}{0 0 533 168}
& \imgcenter[1.5cm]{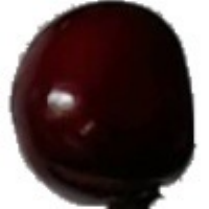}{0 0 101 101}
&  The cluster consists of red-colored fruits. The candidate fruit is also red, though darker in tone. Based on color similarity, it should be included. The answer is \texttt{<CONCLUSION>} \textcolor{red}{\textbf{YES}} \texttt{</CONCLUSION>}.\\
\hline
\end{tabular}
}
\label{tab:appendix1}
\end{table}

\begin{table}[t]
\caption{Examples of Agent-Based Cluster Merge Assessment.}
\centering
\small
\renewcommand\arraystretch{1.6}
\resizebox{1.0\linewidth}{!}{
\begin{tabular}{|>{\centering\arraybackslash}m{1cm}|m{3.5cm}|>{\centering\arraybackslash}m{3.5cm}|>{\centering\arraybackslash}m{3.5cm}|m{3.5cm}|}
\hline
\textbf{Aspect} & \textbf{Prompt} & \textbf{Cluster Centroids} & \textbf{Candidate Centroids} & \textbf{MLLM Generation} \\
\hline
\multirow{3}{*}[-25mm]{\centering Number} 
& \multirow{3}{=}[-15mm]{\justifying Determine whether the two playing card clusters should be merged based on \textcolor{red}{\textbf{rank}}. Ignore the suits and focus only on rank comparison. Respond with \texttt{<CONCLUSION> YES </CONCLUSION>} or \texttt{<CONCLUSION> NO </CONCLUSION>}.}
& \imgcenter[3.5cm]{fig/cluster_centroids1.pdf}{0 0 729 240}
& \imgcenter[3.5cm]{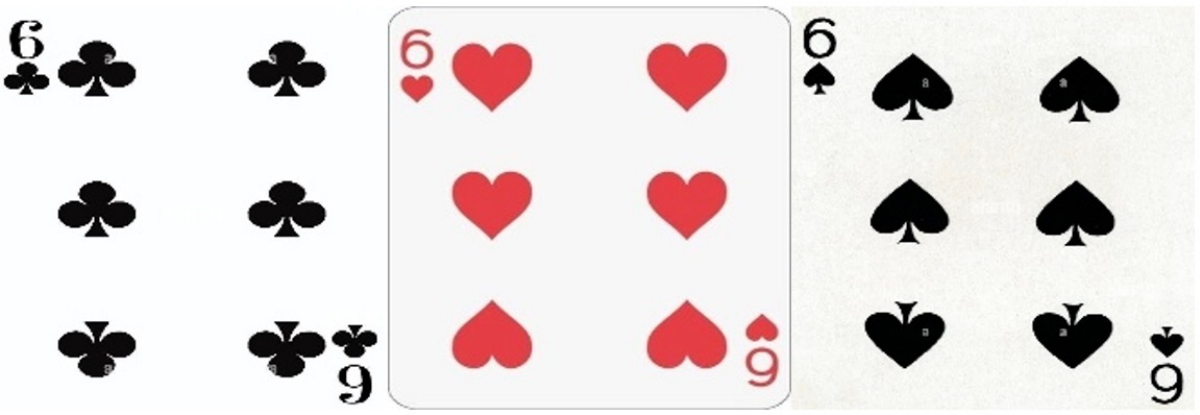}{0 0 576 200}
& The cluster centroids contain cards of rank 9. The candidate centroids consist of cards of rank 6. Since the ranks differ, the clusters should not be merged. The answer is \texttt{<CONCLUSION>} \textcolor{red}{\textbf{NO}} \texttt{</CONCLUSION>}.\\

\cline{3-5}
& 
& \imgcenter[3.5cm]{fig/cluster_centroids2.pdf}{0 0 704 242}
& \imgcenter[3.5cm]{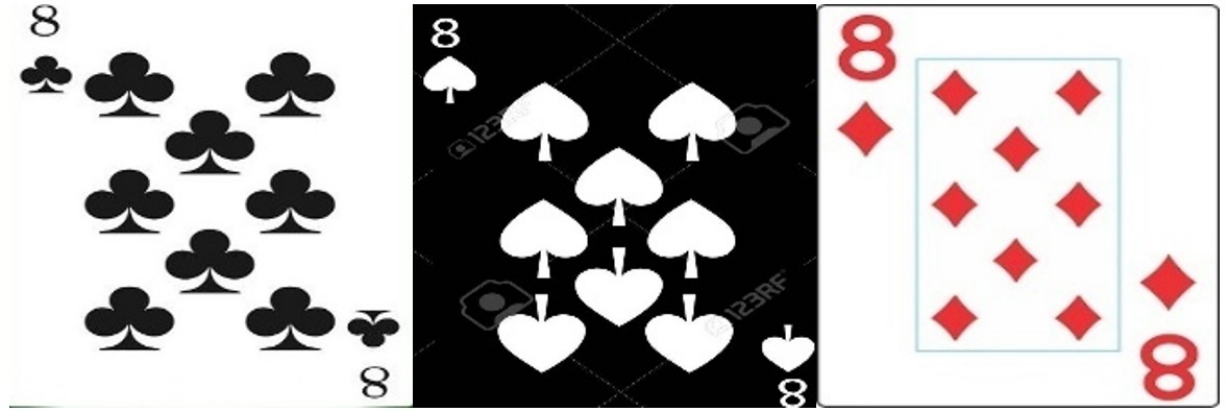}{0 0 589 197}
& The cluster centroids consist of cards ranked 3. The candidate centroids consist of cards ranked 8. Due to the rank difference, the clusters should not be merged. The answer is \texttt{<CONCLUSION>} \textcolor{red}{\textbf{NO}} \texttt{</CONCLUSION>}.\\

\cline{3-5}
& 
& \imgcenter[3.5cm]{fig/cluster_centroids3.pdf}{0 0 720 240}
& \imgcenter[3.5cm]{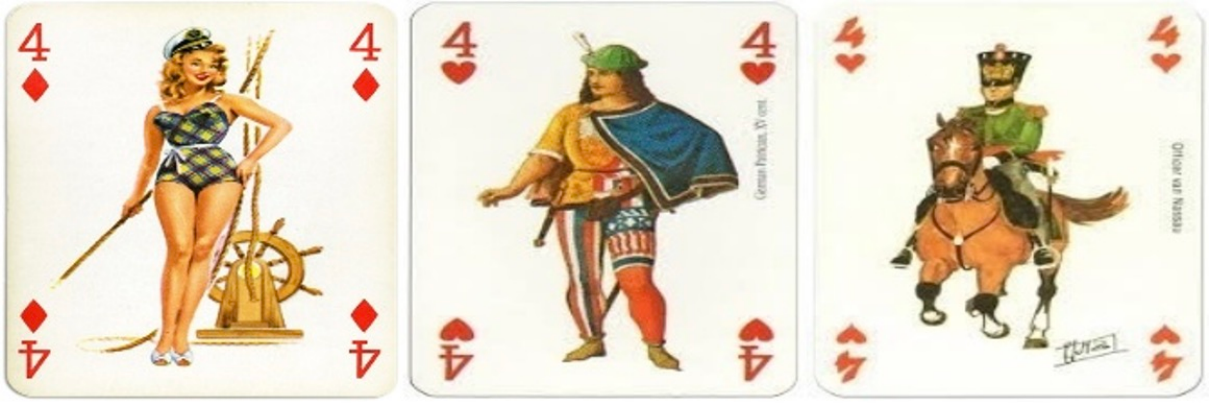}{0 0 580 195}
& Both cluster and candidate centroids consist of cards ranked 4. Based on rank similarity, the clusters should be merged. The answer is \texttt{<CONCLUSION>} \textcolor{red}{\textbf{YES}} \texttt{</CONCLUSION>}.\\
\hline

\multirow{3}{*}[-25mm]{\centering Suits} 
& \multirow{3}{=}[-10mm]{\justifying Determine whether the two playing card clusters should be merged based on \textcolor{red}{\textbf{suits}}. Ignore the rank and focus only on suits comparison. Respond with \texttt{<CONCLUSION> YES </CONCLUSION>} or \texttt{<CONCLUSION> NO </CONCLUSION>}.}
& \imgcenter[3.5cm]{fig/cluster_centroids4.pdf}{0 0 587 195}
& \imgcenter[3.5cm]{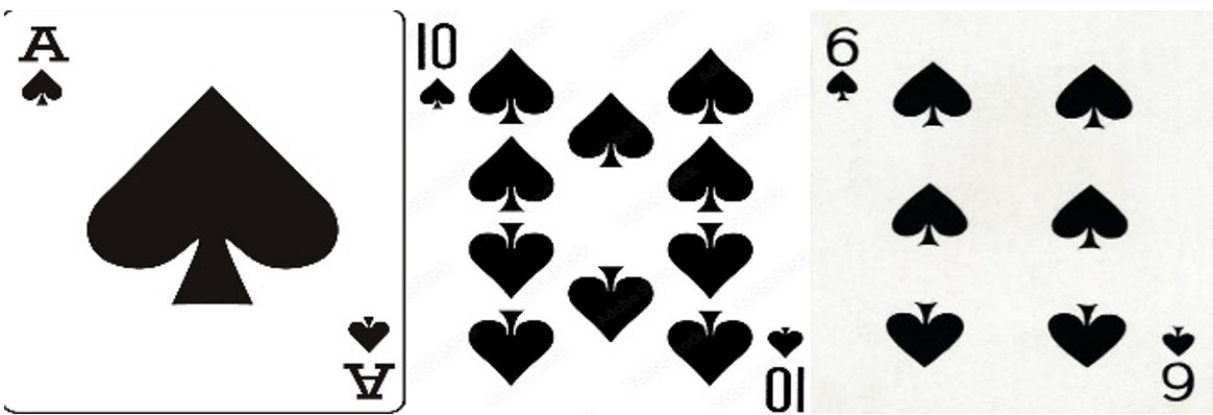}{0 0 582 200}
& The cluster centroids consist of cards from the club suit, while the candidate centroids consist of spade suit cards. As the suits differ, the clusters should not be merged. The answer is \texttt{<CONCLUSION>} \textcolor{red}{\textbf{NO}} \texttt{</CONCLUSION>}.\\

\cline{3-5}
& 
& \imgcenter[3.5cm]{fig/cluster_centroids5.pdf}{0 0 724 243}
& \imgcenter[3.5cm]{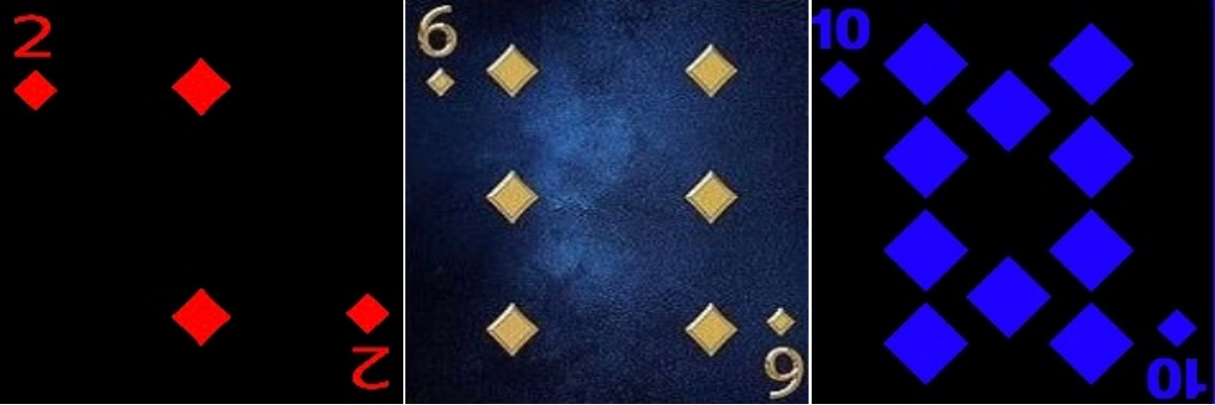}{0 0 584 195}
& Both the cluster and candidate centroids consist of cards from the diamond suit. As the suits match, the clusters should be merged. The answer is \texttt{<CONCLUSION>} \textcolor{red}{\textbf{YES}} \texttt{</CONCLUSION>}.\\

\cline{3-5}
& 
& \imgcenter[3.5cm]{fig/cluster_centroids6.pdf}{0 0 717 242}
& \imgcenter[3.5cm]{fig/cluster_centroids13.pdf}{0 0 582 200}
& The cluster centroids consist of cards from the heart suit, while the candidate centroids consist of cards from the spade suit. Due to suit difference, the clusters should not be merged. The answer is \texttt{<CONCLUSION>} \textcolor{red}{\textbf{NO}} \texttt{</CONCLUSION>}.\\
\hline

\multirow{2}{*}[-12mm]{\centering Color} 
& \multirow{2}{=}[0mm]{\justifying Determine whether the two fruit clusters should be merged based on \textcolor{red}{\textbf{color}}
similarity. Ignore the species and focus only on color comparison. Respond with \texttt{<CONCLUSION> YES </CONCLUSION>}, or \texttt{<CONCLUSION> NO </CONCLUSION>}.}
& \imgcenter[3.5cm]{fig/cluster_centroids7.pdf}{0 0 548 170}
& \imgcenter[3.5cm]{fig/cluster_centroids8.pdf}{0 0 546 166}
& The cluster centroids consist of yellow-colored fruits, while the candidate centroids consist of green-colored fruits. Due to the color difference, the clusters should not be merged. The answer is \texttt{<CONCLUSION>} \textcolor{red}{\textbf{NO}} \texttt{</CONCLUSION>}. \\

\cline{3-5}
& 
& \imgcenter[3.5cm]{fig/cluster_centroids9.pdf}{0 0 533 168}
& \imgcenter[3.5cm]{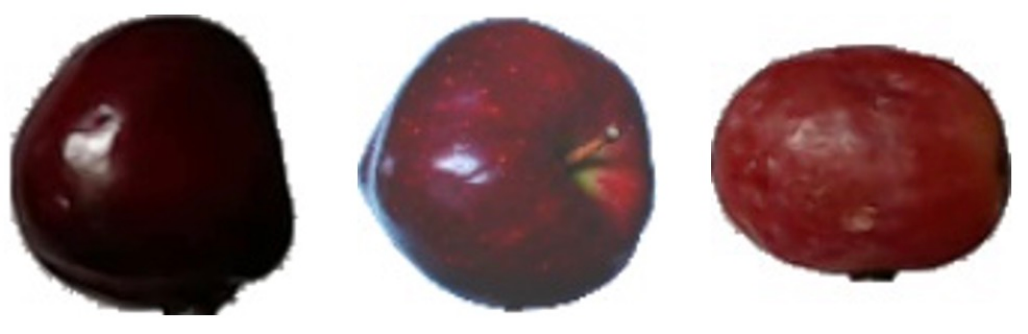}{0 0 494 161}
& Both the cluster and candidate centroids consist of red-colored fruits. As the colors are consistent, the clusters should be merged. The answer is \texttt{<CONCLUSION>} \textcolor{red}{\textbf{YES}} \texttt{</CONCLUSION>}.\\
\hline
\end{tabular}
}
\label{tab:appendix2}
\end{table}

\end{document}